\RequirePackage{booktabs}

\documentclass[pdflatex,sn-mathphys-num]{sn-jnl}

\usepackage{graphicx}%
\usepackage{multirow}%
\usepackage{amsmath,amssymb,amsfonts}%
\usepackage{amsthm}%
\usepackage{mathrsfs}%
\usepackage[title]{appendix}%
\usepackage{xcolor}%
\usepackage{textcomp}%
\usepackage{manyfoot}%
\usepackage{booktabs}%
\usepackage{algorithm}%
\usepackage{algorithmicx}%
\usepackage{algpseudocode}%
\usepackage{listings}%

\def\hd{HD}
\def\hdlong{Hausdorff distance}

\def\hdpercp{HD$_{\!p}$} 
\newcommand{\hdpercn}[1]{HD\textsubscript{#1}} 

\def\masd{MASD}
\def\masdlong{mean average surface distance}
\def\Masdlong{Mean average surface distance}
\def\assd{ASSD}
\def\assdlong{average symmetric surface distance}
\def\Assdlong{Average symmetric surface distance}
\def\nsd{NSD}
\def\nsdlong{normalized surface distance}
\def\Nsdlong{Normalized surface distance}
\def\biou{BIoU}
\def\bioulong{boundary intersection over union}
\def\Bioulong{Boundary intersection over union}
\def\dsc{DSC}
\def\dsclong{Dice similarity coefficient}

\newcommand{\thspace}{\hspace{0.2em}}
\def\anima{Anima}
\def\animatt{\texttt{\anima}}
\def\evaluatesegmentation{EvaluateSegmentation}
\def\evaluatesegmentationtt{\texttt{\evaluatesegmentation}}
\def\googledeepmind{Google\thspace DeepMind}
\def\googledeepmindtt{\texttt{\googledeepmind}}
\def\medpy{MedPy}
\def\medpytt{\texttt{\medpy}}
\def\metricsreloaded{Metrics\thspace Reloaded}
\def\metricsreloadedtt{\texttt{\metricsreloaded}}
\def\metricsreloadedit{\textit{\metricsreloaded}}
\def\miseval{MISeval}
\def\misevaltt{\texttt{\miseval}}
\def\monai{MONAI}
\def\monaitt{\texttt{\monai}}
\def\plastimatch{Plastimatch}
\def\plastimatchtt{\texttt{\plastimatch}}
\def\pymia{pymia}
\def\pymiatt{\texttt{\pymia}}
\def\segmetrics{seg-metrics}
\def\segmetricstt{\texttt{\segmetrics}}
\def\simpleitk{SimpleITK}
\def\simpleitktt{\texttt{\simpleitk}}

\def\segA{$A$}
\def\segAmask{$A_\mathcal{M}$}

\def\msegApoints{A_\mathcal{P}}
\def\segAboundary{$A_\mathcal{B}$}
\def\msegAboundary{A_\mathcal{B}}

\def\segB{$B$}
\def\segBmask{$B_\mathcal{M}$}

\def\msegBpoints{B_\mathcal{P}}
\def\segBboundary{$B_\mathcal{B}$}
\def\msegBboundary{B_\mathcal{B}}

\def\PosDec{\texttt{$\mathbb R_{>0}$}} 

\def\spacingOneOne{$(1.0{,}\,1.0)$\,mm}
\def\spacingPointZeroSevenPointZeroSeven{$(0.07{,}\,0.07)$\,mm}
\def\spacingPointZeroSevenOne{$(0.07{,}\,1.0)$\,mm}

\def\spacingOneOneOne{$(1.0{,}\,1.0{,}\,1.0)$\,mm}

\def\spacingTwoTwoTwo{$(2.0{,}\,2.0{,}\,2.0)$\,mm}

\def\spacingHalfHalfTwo{$(0.5{,}\,0.5{,}\,2.0)$\,mm}

\usepackage{xstring}
\usepackage{array}
\usepackage{tabularx}
\usepackage{hhline}
\usepackage{graphbox} 
\usepackage{pifont}
	\newcommand{\cmark}{\ding{51}}%
	\newcommand{\xmark}{\ding{55}}%

\usepackage{dcolumn}
\usepackage{colortbl}
\usepackage{changepage} 
\usepackage{hyperref}
	\hypersetup{colorlinks,	citecolor=blue,	filecolor=blue,	linkcolor=blue,	urlcolor=blue}
\usepackage[margin=0pt,font={small},labelfont={bf},labelsep=period]{caption}
\usepackage{subcaption}
\usepackage{soul}
\usepackage{geometry}
\usepackage{natbib}
\usepackage{titlesec}
	\titleformat{\subsection}{\normalfont\fontsize{12}{14}\bfseries}{\thesubsection}{1em}{}
	\titleformat{\subsubsection}[runin]{\bfseries}{}{}{}[\hspace*{0.8em}]
	\titlespacing*{\section}{0pt}{16pt plus 4pt minus 2pt}{5pt plus 2pt minus 2pt}
	\titlespacing*{\subsection}{0pt}{12pt plus 4pt minus 2pt}{5pt plus 2pt minus 2pt}
    \titlespacing*{\subsubsection}{0pt}{9pt plus 4pt minus 2pt}{2pt plus 2pt minus 2pt}   
\usepackage{makecell}
\usepackage{setspace}
\usepackage{adjustbox}
\makeatletter
\newcommand\footnoteref[1]{\protected@xdef\@thefnmark{\ref{#1}}\@footnotemark}
\makeatother
\begin{document}
\title[Distance-based metrics implementation pitfalls]{\centering \textbf{Understanding implementation pitfalls of distance-based metrics for image segmentation}}
\author[*]{\fnm{Ga\v{s}per} \sur{Podobnik}}
\author{\fnm{Toma\v{z}} \sur{Vrtovec}}
\affil{\normalsize
	\orgname{University of Ljubljana},
	\orgdiv{Faculty of Electrical Engineering},  \orgaddress{\street{Tr\v{z}a\v{s}ka cesta 25}, \postcode{SI-1000} \city{Ljubljana}, \country{Slovenia}}}
\abstract{%
Distance-based metrics, such as the Hausdorff distance (\hd), are widely used to validate segmentation performance in (bio)medical imaging. However, their implementation is complex, and critical differences across open-source tools remain largely unrecognized by the community. These discrepancies undermine benchmarking efforts, introduce bias in biomarker calculations, and potentially distort medical device development and clinical commissioning. In this study, we systematically dissect 11 open-source tools that implement distance-based metric computation by performing both a \textit{conceptual analysis} of their computational steps and an \textit{empirical analysis} on representative two- and three-dimensional image datasets. Alarmingly, we observed deviations in \hd\ exceeding 100\,mm and identified multiple statistically significant differences between tools -- demonstrating that statistically significant improvements on the same set of segmentations can be achieved simply by selecting a particular implementation. These findings cast doubts on the validity of prior comparisons of results across studies without accounting for the differences in metric implementations. To address this, we provide practical recommendations for tool selection; additionally, our conceptual analysis informs about the future evolution of implementing open-source tools.}
\maketitle
\section{Introduction}
Method validation is the process of assessing how well a method performs in achieving its intended objectives through the analysis of its accuracy, efficiency and reliability under specific conditions. In the field of image analysis spanning applications in (bio)medical imaging, performance evaluation constitutes a cornerstone of every study~\cite{Maier-Hein2024-Metrics-Reloaded, Margolin2014-Evaluate-Foreground, BeyondRoss2023-BeyondRankings}. Metrics serve as proxies for evaluating task performance, and are broadly categorized into \textit{qualitative} and \textit{quantitative}. Qualitative metrics, commonly based on visual inspection, offer focused and direct evaluations, however, they often rely on subjective expert judgments, rendering them costly and occasionally impractical or even unfeasible. In contrast, quantitative metrics are swiftly computed, theoretically well-defined, and thus notably objective. While a single metric\footnote{In the context of this study, \textit{quantitative metrics} are hereinafter referred to simply as \textit{metrics}.} may not always provide a holistic performance measure, a combination of multiple metrics can provide a solid foundation for facilitating decision-making processes, rapid method prototyping, ablation studies, benchmarking and, generally, the development of novel methodologies. 
\par
Image segmentation is a fundamental task in (bio)medical image analysis~\cite{Antonelli2022_Decathlon, Ma2024_SegmentAnything} for which a plethora of validation metrics have been proposed. These include the well-established ones, such as the \textit{intersection over union} (IoU), \textit{\dsclong} (\dsc), \textit{\hdlong} (\hd) with its $p$-th percentile variants (\hdpercp), \textit{\masdlong} (\masd) and \textit{\assdlong} (\assd), as well as the more recently proposed ones, such as the \textit{\nsdlong} (\nsd)~\cite{Nikolov2021-DeepMind} and \textit{\bioulong} (\biou)~\cite{Cheng2021BoundaryIoU}. Given the multitude of the proposed metrics, careful \textit{metric selection} is essential to ensure that the reported segmentation performance aligns with clinical objectives. The importance of this step is illustrated by several guidelines for metric selection proposed in the literature~\cite{Taha2015-EvaluateSegmentation, Muller2022-MISeval, Maier-Hein2024-Metrics-Reloaded}. Once selected, metrics must be also correctly implemented, i.e.\ their mathematical definitions must be faithfully translated into computational code. Incorrect implementations can have significant ramifications, including flawed benchmarking, biased validation, distorted method development, and compromised medical device commissioning~\cite{Reinke2024-Metrics-Pitfalls, Maier2018Whyrankings}. This concern is amplified in the modern landscape of rapid research dissemination, where results are often compared against previously reported studies rather than reimplemented baselines, under the implicit assumption that metric implementations are consistent and equivalent across the community. Currently, there is no single standardized open-source library that encompasses all segmentation metrics. Instead, countless in-house and several open-source tools are adopted in specific domains of (bio)medical image analysis~\cite{Commowick2018Anima, Taha2015-EvaluateSegmentation, Nikolov2021-DeepMind, Maier2019MedPy, Maier-Hein2024-Metrics-Reloaded, Muller2022-MISeval, Cardoso2022-MONAI, Zaffino2016-Plastimatch, Jungo2021pymia, Jia2024-SegMetrics, Lowekamp2013SimpleITK}. To date, the validity and consistency of these tools have not been systematically scrutinized. Our preliminary study revealed substantial discrepancies in \hd\ implementations, resulting in large deviations in metric values~\cite{Podobnik2024HDilemma}. These findings underscore an acute problem that has yet to be adequately addressed, and motivate a comprehensive analysis of implementation pitfalls and a forward-looking discussion of potential solutions.
\subsection{Related works}
The closest relevant works to the topic of metric implementation pitfalls are metric selection guidelines~\cite{Taha2015-EvaluateSegmentation, Muller2022-MISeval, Maier-Hein2024-Metrics-Reloaded, DrukkerMIDRC2024}. One of such early endeavors is the correlation analysis between different metrics by Taha and Hanbury~\cite{Taha2015-EvaluateSegmentation} that highlighted the importance of distance-based metrics, particularly \hd, and proposed an efficient implementation of \hd~\cite{Taha2015-HD-Optimization}. A related investigation by Müller et al.~\cite{Muller2022-MISeval} identified several common pitfalls in the research community, including \textit{incorrect metric implementation}, although this pitfall was not systematically analyzed. They proposed several guidelines, notably advocating for open access to evaluation scripts via open-source platforms\footnote{\href{https://github.com/}{GitHub}, \href{https://about.gitlab.com/}{GitLab} or \href{https://zenodo.org/}{Zenodo}.}. The most comprehensive effort to date on metric selection guidelines for various medical imaging tasks -- including classification, detection, and semantic/instance segmentation -- was published by Maier-Hein et al.~\cite{Maier-Hein2024-Metrics-Reloaded, Reinke2024-Metrics-Pitfalls}. 
This collaborative consortium of medical imaging researchers identified key pitfalls in problem categorization, metric selection and application~\cite{Reinke2024-Metrics-Pitfalls}, culminating in the release of the \metricsreloadedit\ guidelines~\cite{Maier-Hein2024-Metrics-Reloaded}. Among the pitfalls, \textit{inadequate metric implementation} was noted, particularly inconsistencies in how segmentation boundaries are extracted, which is a crucial computation step for all distance-based metrics. These variations across tools can affect metric values, yet no specific guidance was provided for resolving such discrepancies.
\subsection{Purpose and contributions}
Given the central role of validation metrics in (bio)medical image analysis, the variability in distance-based metric implementations represents a critical yet underexplored issue. To systematically assess how such inconsistencies affect the validation process, we undertake a twofold analysis (Fig.~\ref{fig:experimental-design}). First, a \textit{conceptual analysis} compares boundary extraction methods, mathematical definitions and edge case handling to pinpoint the sources of variability. Second, an \textit{empirical analysis} evaluates differences in metric values and computational efficiency across tools on real clinical datasets, quantifying the impact of implementation pitfalls. Our study focuses on distance-based metrics -- including \hd, \hdpercp, \masd, \assd, \nsd\ and \biou\ -- which are widely used and share common computational principles. This subset aligns with the one in the \metricsreloadedit\ guidelines~\cite{Maier-Hein2024-Metrics-Reloaded}, and provides a coherent framework for a comprehensive assessment.
\phantomsection
\section{Methods}\label{sec:methods}
\subsection{Open-source tools}
We conducted a web search for open-source tools that provide distance-based metric computation with the following inclusion criteria: (1)~the tool implements at least one of the distance-based metrics (i.e.\ \hd, \hdpercp, \masd, \assd, \nsd\ or \biou), (2)~the tool supports two-dimensional (2D) as well as three-dimensional (3D) computations, and (3)~the tool has a command-line or \texttt{Python} interface. 
\begin{figure}[!t]
	\centering
	\includegraphics[width=0.45\textwidth]{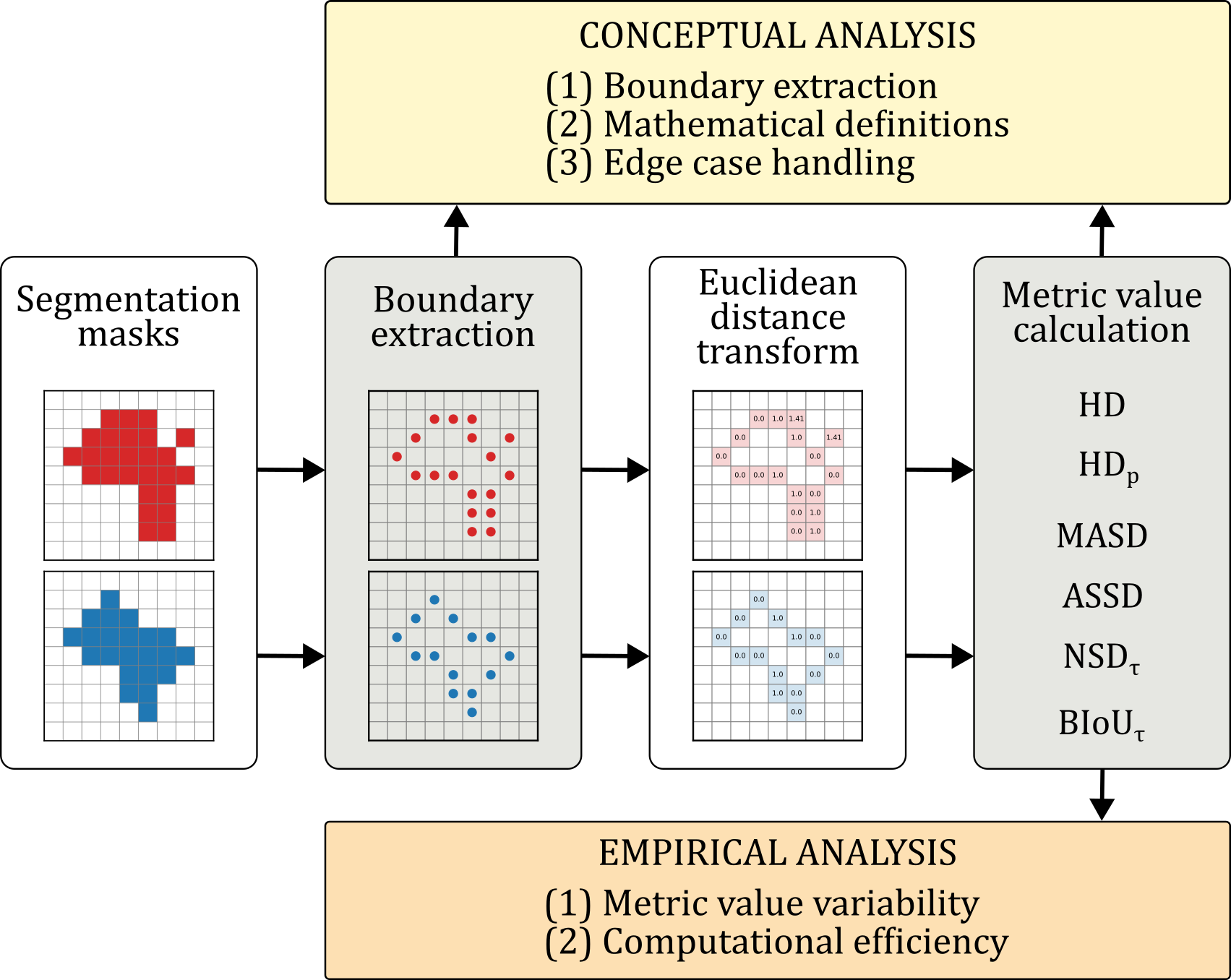}
	\vspace{0.25cm}
	\caption{A schematic depiction of the study design and the distance-based metric calculation workflow.}
	\label{fig:experimental-design}
\end{figure}
\subsection{Conceptual analysis}
To contextualize the conceptual analysis, we first provide a brief overview of the distance-based metric computation workflow, also illustrated in Fig.~\ref{fig:experimental-design}. Starting from a pair of segmentations, e.g.\ a reference and a prediction, the process begins with extracting their respective boundaries. These boundaries are then used to compute two sets of directed distances: from the reference to the prediction, and vice versa. Finally, a metric-specific aggregation function (based on a mathematical definition) transforms these distance sets into a single scalar, i.e.\ the metric value. According to our preliminary investigation~\cite{Podobnik2024HDilemma}, we identified three primary sources of variability in this workflow, which represent the focus of this conceptual analysis: (1)~boundary extraction algorithms, (2)~variations in the underlying mathematical definitions, and (3)~edge case handling (i.e.\ metric value when one or both segmentations are empty).
\subsection{Empirical analysis}
Building on the findings from the conceptual analysis, the empirical analysis seeks to quantify the impact of implementation pitfalls on both metric values and the computational efficiency of open-source tools. To account for both 2D and 3D segmentation use cases, we conduct a comprehensive assessment using one publicly available dataset of each type. For the 2D analysis, we selected the \textit{INbreast dataset}~\cite{Moreira2012INbreast}, which includes high-resolution mammograms from 115 cases with corresponding manual tumor segmentations. An additional set of segmentations was generated using an in-house method, resulting in $80$ pairs of 2D segmentation masks. For the 3D analysis, we used a subset of the \textit{HaN-Seg dataset}~\cite{Podobnik2023-Han-Seg-Dataset}, comprising $30$ computed tomography and magnetic resonance image pairs from the radiotherapy planning workflow. Each image was paired with 3D segmentations of up to $30$ organs-at-risk, independently delineated by two clinical experts, yielding 1,559 pairs of 3D segmentation masks.
\par
In contrast to counting-based metrics, distance-based metrics rely on spatial computations of distances, and therefore require image elements to be defined in physical units, i.e.\ with a specified pixel size $(d_x, d_y)$ in 2D or voxel size $(d_x, d_y, d_z)$ in 3D.  Proper handling of image element sizes is essential for correct metric implementations. To assess this, we evaluate three different element sizes for each dataset: two isotropic (a \textit{vanilla scenario} with unit size and a non-unit size) and one anisotropic. For the 2D analysis, we considered: (1)~\spacingOneOne, (2)~\spacingPointZeroSevenPointZeroSeven, representing a common isotropic pixel size in mammography, and (3)~\spacingPointZeroSevenOne. For the 3D analysis, we used: (1)~\spacingOneOneOne, (2)~\spacingTwoTwoTwo, and (3)~\spacingHalfHalfTwo\ -- all of which represent voxel sizes typical of radiotherapy workflows and thus reflect the evaluation settings commonly encountered in practice.
\par
The empirical analysis involves loading the original segmentations at their native resolution, resampling them to each of the three pixel/voxel sizes, and computing the distance-based metrics using all identified open-source tools. For metrics requiring user-defined parameters, i.e.\ the percentile $p$ for \hdpercp\ and the boundary margin of error $\tau$ for \nsd\ and \biou, we performed experiments with \hdpercn{\!$p$\,{=}\,95}, \nsd$_{\tau\,{=}\,2\,\textnormal{mm}}$ and \biou$_{\tau\,{=}\,2\,\textnormal{mm}}$, which are commonly used parameter values~\cite{Maier-Hein2024-Metrics-Reloaded}.
\phantomsection
\section{Results}
\begin{table*}[!t]
	\centering
	\caption{Overview of $11$ open-source tools analyzed in this study, indicating the supported distance-based metrics (names are hyperlinked to corresponding publications, commit hashes are hyperlinked to publicly available code repositories). For \hdpercp, the checkmark (\checkmark) denotes the support of any percentile, whereas a specific number indicates the implementation of a predefined percentile (e.g.\ 95-th).}
	\vspace{0.25cm}
	\label{tab:implementations-info}
\newcommand{\toolNameDOI}[2]{\href{#2}{\texttt{#1}}}
\newcommand{\toolCommitGitHub}[2]{\href{#2}{\texttt{#1}}}
\def\nameAnima{\toolNameDOI{Anima}{https://doi.org/10.1038/s41598-018-31911-7}}
\def\shaAnima{\toolCommitGitHub{1df7e44}{https://github.com/Inria-Empenn/Anima-Public}}
\def\nameEvaluateSegmentation{\toolNameDOI{EvaluateSegmentation}{https://doi.org/10.1186/s12880-015-0068-x}}
\def\shaEvaluateSegmentation{\toolCommitGitHub{4cff08d}{https://github.com/Visceral-Project/EvaluateSegmentation}}
\def\nameDeepMind{\toolNameDOI{Google DeepMind}{https://doi.org/10.2196/26151}}
\def\shaDeepMind{\toolCommitGitHub{1f805ce}{https://github.com/google-deepmind/surface-distance}}
\def\nameMedPy{\toolNameDOI{MedPy}{https://doi.org/10.5281/zenodo.2565940}}
\def\shaMedPy{\toolCommitGitHub{d6abf41}{https://github.com/loli/medpy/}}
\def\nameMetricsReloaded{\toolNameDOI{Metrics Reloaded}{https://doi.org/10.1038/s41592-023-02151-z}}
\def\shaMetricsReloaded{\toolCommitGitHub{cb38dfc}{https://github.com/Project-MONAI/MetricsReloaded}}
\def\nameMISeval{\toolNameDOI{MISeval}{https://doi.org/10.3233/shti220391}}
\def\shaMISeval{\toolCommitGitHub{2aa659f}{https://github.com/frankkramer-lab/miseval}}
\def\nameMONAI{\toolNameDOI{MONAI}{https://doi.org/10.48550/arXiv.2211.02701}}
\def\shaMONAI{\toolCommitGitHub{d388d1c}{https://github.com/Project-MONAI/MONAI}}
\def\namePlastimatch{\toolNameDOI{Plastimatch}{https://doi.org/10.1118/1.4961121}}
\def\shaPlastimatch{\toolCommitGitHub{785864a7}{https://gitlab.com/plastimatch/plastimatch}}
\def\namepymia{\toolNameDOI{pymia}{https://doi.org/10.1016/j.cmpb.2020.105796}}
\def\shapymia{\toolCommitGitHub{9afc15f}{https://github.com/rundherum/pymia}}
\def\nameSegMetrics{\toolNameDOI{seg-metrics}{https://doi.org/10.1101/2024.02.22.24303215}}
\def\shaSegMetrics{\toolCommitGitHub{9513586}{https://github.com/Jingnan-Jia/segmentation_metrics}}
\def\nameSimpleITK{\toolNameDOI{SimpleITK}{https://doi.org/10.3389/fninf.2013.00045}}
\def\shaSimpleITK{\toolCommitGitHub{2018b6b}{https://github.com/SimpleITK/SimpleITK}}
\def\shaMeshMetrics{\toolCommitGitHub{e1a0455}{https://github.com/gasperpodobnik/MeshMetrics}}
\newcommand{\LL}[2]{%
    \IfStrEq{#2}{S}{\textsuperscript{#1}}{}%
    \IfStrEq{#2}{D}{%
        \IfEqCase{#1}{%
            {a}{\animatt: Metric name is \texttt{ContourMeanDistance} for \masd, and \texttt{SurfaceDistance} for \assd.}%
            {b}{\evaluatesegmentationtt: Metric name is \texttt{ASD} for \masd.}%
            {c}{\googledeepmindtt: Function \texttt{compute\_average\_surface\_distance} returns the two directed average distances and we take their mean to obtain \masd, while function \texttt{compute\_surface\_distances} returns vectors of distances and corresponding boundary element sizes that we use to calculate \assd.}%
            {d}{\misevaltt: Function \texttt{calc\_SimpleHausdorffDistance} only works for 2D, therefore we use the more general \texttt{calc\_AverageHausdorffDistance} to calculate \hd\ both in 2D and 3D, which is the implementation of \hdpercn{100} despite its name.}%
            {e}{\monaitt: We use \texttt{compute\_average\_surface\_distance} with argument \texttt{symmetric} set to \texttt{True} to calculate \assd.}%
            {f}{\plastimatchtt: Metric name is \texttt{Hausdorff distance (boundary)} for \hd, \texttt{Percent (0.95) Hausdorff distance (boundary)} for \hdpercn{95}, and \texttt{Avg average Hausdorff distance (boundary)} for \masd.}%
            {g}{\pymiatt: Metric name is \texttt{AverageDistance} for \masd.}%
            {h}{\segmetricstt: Metric name is \texttt{msd} for \assd.}%
            {i}{\simpleitktt: We use method \texttt{GetAverageHausdorffDistance} of the \texttt{HausdorffDistanceImageFilter} class to calculate \masd.}%
        }%
    }{}%
}

\def\cpp{\texttt{C++}}
\def\python{\texttt{Python}}
\def\dims{\checkmark\checkmark}

\setlength{\tabcolsep}{4.45pt}
\renewcommand{\arraystretch}{1.3}
\small
\newcommand{\rotC}[1]{\rotatebox{90}{#1}}
\newcolumntype{x}[1]{>{\arraybackslash\hspace{0pt}}p{#1}}
\newcommand{\addFnote}[1]{\multicolumn{11}{x{0.985\textwidth}}{\footnotesize{#1}}}

\newcolumntype{L}[1]{>{\raggedright\let\newline\\\arraybackslash\hspace{0pt}}m{#1}}
\newcolumntype{C}[1]{>{\centering\let\newline\\\arraybackslash\hspace{0pt}}m{#1}}
\newcolumntype{R}[1]{>{\raggedleft\let\newline\\\arraybackslash\hspace{0pt}}m{#1}}
\def\clen{1.0cm}

\begin{tabularx}{\linewidth}{L{4.2cm}ccL{1.4cm} X *{6}{C{\clen}}}
\hline \noalign{\smallskip}
& \multicolumn{3}{c}{Details} & & \multicolumn{6}{c}{Distance-based metrics} \\
\cline{2-4} \cline{6-11}
Open-source tool                      & Commit hash              & Version & Language & & \hd    & \hdpercp & \masd           & \assd           & \nsd     & \biou \\ \noalign{\smallskip} \hline \noalign{\smallskip}
\nameAnima~\cite{Commowick2018Anima}                & \shaAnima                & 4.2{~~\,}      & \cpp &     & \checkmark          &                 & \checkmark\LL{a}{S}    & \checkmark\LL{a}{S}    &                 &              \\
\nameEvaluateSegmentation~\cite{Taha2015-EvaluateSegmentation} & \shaEvaluateSegmentation & N/A       & \cpp &     & \checkmark          & 95              & \checkmark\LL{b}{S}    &                        &                 &              \\
\nameDeepMind~\cite{Nikolov2021-DeepMind}             & \shaDeepMind             & 0.1{~~\,}      & \python &  & \checkmark          & \checkmark      & \checkmark\LL{c}{S}    & \checkmark\LL{c}{S}    & \checkmark      &              \\
\nameMedPy~\cite{Maier2019MedPy}                & \shaMedPy                & 0.5.2    & \python &  & \checkmark          & 95              &     &   \checkmark       &                 &              \\
\nameMetricsReloaded~\cite{Maier-Hein2024-Metrics-Reloaded}      & \shaMetricsReloaded      & 0.1.0    & \python &  & \checkmark          & \checkmark      & \checkmark             & \checkmark             & \checkmark      & \checkmark   \\
\nameMISeval~\cite{Muller2022-MISeval}              & \shaMISeval              & 1.3.0    & \python &  & \checkmark\LL{d}{S} &                 &                        &                        &                 &              \\
\nameMONAI~\cite{Cardoso2022-MONAI}                & \shaMONAI                & 1.5.0    & \python &  & \checkmark          & \checkmark      &     & \checkmark\LL{e}{S}    & \checkmark      &              \\
\namePlastimatch~\cite{Zaffino2016-Plastimatch}          & \shaPlastimatch          & 1.10.0    & \cpp &     & \checkmark\LL{f}{S} & 95\LL{f}{S}     & \checkmark\LL{f}{S}    &                        &                 &              \\
\namepymia~\cite{Jungo2021pymia}                & \shapymia                & 0.3.4    & \python &  & \checkmark          & \checkmark      & \checkmark\LL{g}{S}    &                        & \checkmark      &              \\
\nameSegMetrics~\cite{Jia2024-SegMetrics}           & \shaSegMetrics           & 1.2.8    & \python &  & \checkmark          & 95              &                        & \checkmark\LL{h}{S}    &                 &              \\
\nameSimpleITK~\cite{Lowekamp2013SimpleITK}            & \shaSimpleITK            & 2.5.2    & \python &  & \checkmark          &                 & \checkmark\LL{i}{S}    &                        &                 &              \\ 
\noalign{\smallskip} \hline \noalign{\smallskip}
\addFnote{%
    \LL{a\,}{S}\LL{a}{D}\ %
    \LL{b\,}{S}\LL{b}{D}\ %
    \LL{c\,}{S}\LL{c}{D}\ %
    \LL{d\,}{S}\LL{d}{D}\ %
    \LL{e\,}{S}\LL{e}{D}\ %
    \LL{f\,}{S}\LL{f}{D}\ %
    \LL{g\,}{S}\LL{g}{D}\ %
    \LL{h\,}{S}\LL{h}{D}\ %
    \LL{i\,}{S}\LL{i}{D}\ %
}\\
\noalign{\smallskip} \hline
\end{tabularx}
\end{table*}
\subsection{Open-source tools}
The search resulted in the following 11 open-source tools (listed in alphabetical order):
\begin{itemize}
	\renewcommand\labelitemi{$\bullet$}
	\setlength\itemsep{0.25em}
	\item \animatt\ -- Initially proposed for the evaluation of multiple sclerosis lesion segmentation method~\cite{Commowick2018Anima}.
	\item \evaluatesegmentationtt\ -- Released by Taha and Hanbury concurrently with their metric selection guidelines~\cite{Taha2015-EvaluateSegmentation}. 
	\item \googledeepmindtt\ -- Introduced alongside the study that proposed \nsd~\cite{Nikolov2021-DeepMind}.
	\item \medpytt\ -- A general-purpose toolbox for medical image processing~\cite{Maier2019MedPy}.
	\item \metricsreloadedtt\ -- Includes implementations of all metrics analyzed within \metricsreloadedit~\cite{Maier-Hein2024-Metrics-Reloaded}.
	\item \misevaltt\ -- Released by Müller et al.\ concurrently with their metric selection guidelines~\cite{Muller2022-MISeval}.
	\item \monaitt\ -- Widely recognized by the (bio)medical image analysis community and supported by NVIDIA~\cite{Cardoso2022-MONAI}.
	\item \plastimatchtt\ -- Established for biomedical image registration and segmentation~\cite{Zaffino2016-Plastimatch}, utilized within \textit{SlicerRT}, a radiotherapy extension of the open-source software \textit{3D Slicer}~\cite{Fedorov2012_3DSlicer}.
	\item \pymiatt\ -- Developed for data handling and evaluation in (bio)medical image analysis~\cite{Jungo2021pymia}.
	\item \segmetricstt\ -- Recently released for segmentation metric calculation~\cite{Jia2024-SegMetrics}.
	\item \simpleitktt\ -- Multi-dimensional image analysis based on \textit{Insight Toolkit} (ITK)~\cite{Lowekamp2013SimpleITK}.
\end{itemize}
Table~\ref{tab:implementations-info} presents detailed information about all open-source tools, including the programming language they are implemented in and the set of distance-based metrics they support. Additionally, we provide information on the versions and exact commit hashes of the repositories used for our experiments so as to ensure their reproducibility, which correspond to the latest updates for each tool available as of July 2025. The tools differ in various aspects, such as their popularity within the community, inception date, maintenance level, programming language, customization options, as well as naming conventions. Moreover, it has to be noted that their development trajectories are not isolated but rather interconnected, with older tools influencing the design of newer ones. For more details, please refer to Notes~\ref{note:nomenclature} and~\ref{note:similarities} in Appendix.
\subsection{Conceptual analysis}
\subsubsection{Boundary extraction}
A lack of standardization of the boundary extraction is one of the main pitfalls of distance-based metric implementation~\cite{Reinke2024-Metrics-Pitfalls, Maier-Hein2024-Metrics-Reloaded}. Fig.~\ref{fig:boundary-extraction} shows these differences among the tools when probed with two example segmentation masks. The most notable outliers are presented in Fig.~\ref{fig:boundary-extraction-e} and \ref{fig:boundary-extraction-f}, which reveal the foreground- vs.\ boundary-based calculation dilemma. Among the 11 open-source tools, \evaluatesegmentationtt\ and \simpleitktt\ are the only two that omit the boundary extraction step and calculate distances between all foreground elements instead, using the pixel/voxel centers as query points. The authors of \evaluatesegmentationtt\ even proposed several optimization strategies to improve the computational efficiency, such as excluding intersecting elements because their distances are always zero\footnote{\evaluatesegmentationtt\ computes distances solely for non-overlapping elements~\cite{Taha2015-HD-Optimization}, while \simpleitktt\ computes them for all foreground elements.}~\cite{Taha2015-HD-Optimization}. \plastimatchtt\ is the only tool that returns both foreground- and boundary-based calculations, while other tools support boundary-based calculations only. Among these, the most frequently employed boundary extraction method is utilized by \animatt, \medpytt, \metricsreloadedtt, \misevaltt, \monaitt\ and \plastimatchtt, and involves morphological erosion using a square-connectivity structural element, while \segmetricstt\ uses a full-connectivity structural element. Conversely, \googledeepmindtt\ and \pymiatt\ employ a strategy where the image grid is shifted by half a pixel/voxel size. These two implementations are the only two that also calculate boundary element sizes\footnote{The lengths of line segments in 2D and areas of surface elements in 3D.} and use them in metric calculation as noted in the analysis of the mathematical definitions.
\begin{figure*}[!t]
	\centering
	\input{figures/figure_boundary_extraction.tex}
	\vspace{0.25cm}
	\caption{Overview of the boundary extraction methods used by the 11 open-source tools. All methods are demonstrated using 2D examples, but can be seamlessly extended to 3D.}
	\label{fig:boundary-extraction}
\end{figure*}
\subsubsection{Mathematical definitions}
\begin{figure*}[!t]
	\includegraphics{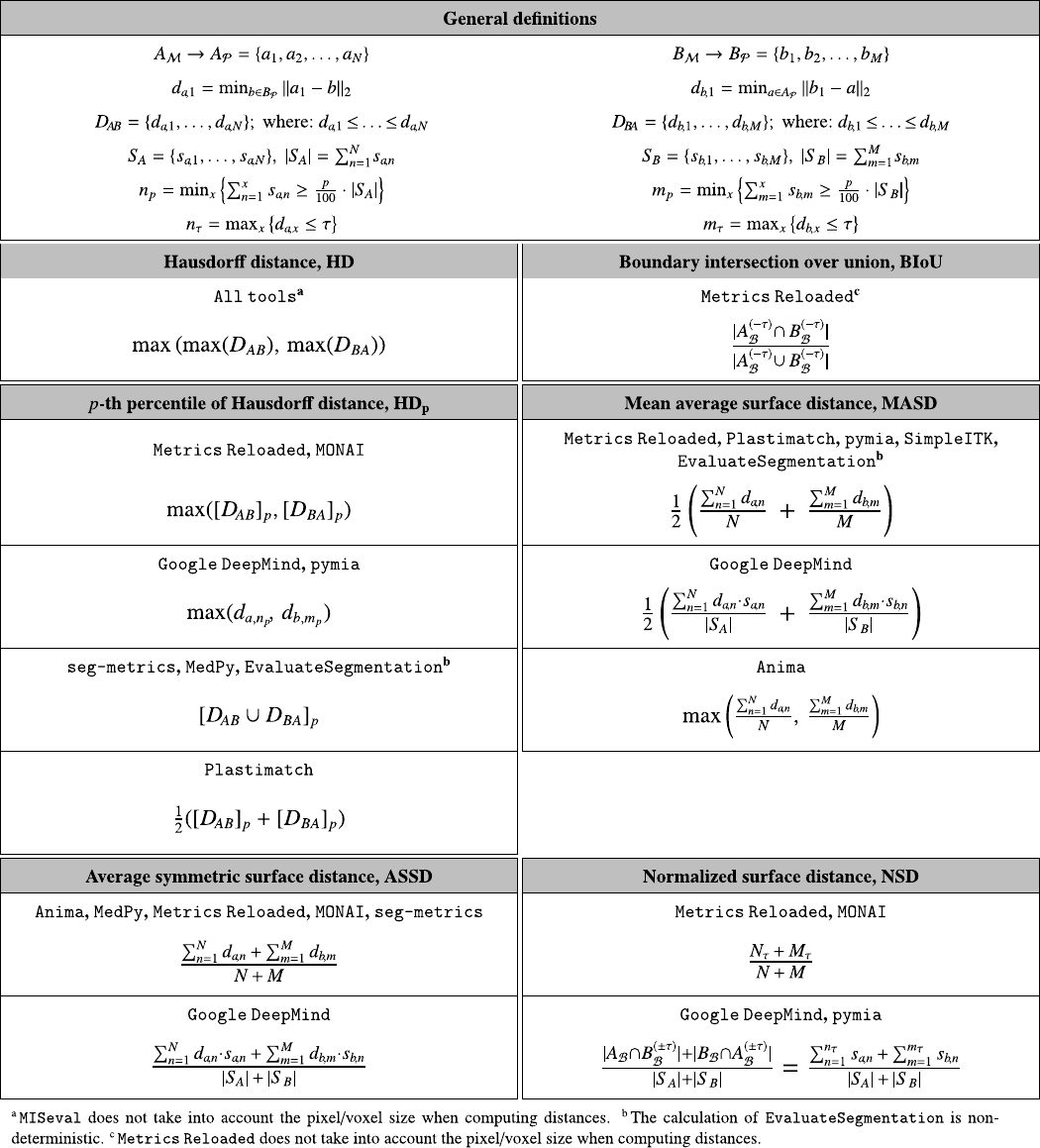}
	\vspace{0.25cm}
	\caption{Mathematical definitions of distance-based metrics as adopted by different open-source tools.}
	\label{fig:equations}
\end{figure*}
In their straightforward application, all distance-based metrics compare two segmentations \segA\ and \segB. All 11 open-source tools support solely calculation on segmentation \textit{masks} \segAmask\ and \segBmask, i.e.\ a grid-based calculation. For simplicity, we treat the segmentations as binary, however, all findings can be generalized to multi-label segmentations. All distance-based metric implementations are based on two finite sets $\msegApoints = \left\{a_1, a_2, \ldots, a_N \right\}$ and $\msegBpoints = \left\{b_1, b_2, \ldots, b_M \right\}$, with cardinalities $N$ and $M$, respectively, which represent point clouds derived from segmentation masks \segAmask\ and \segBmask\ (cf.\ Fig.~\ref{fig:equations}, \textit{General definitions}), and are used to compute distances between these two segmentations. We give a brief historical account, describe the mathematical definition, and present our findings on practical implementation separately for each metric (Fig.~\ref{fig:equations}):
\begin{itemize}
	\renewcommand\labelitemi{$\bullet$}
	\setlength\itemsep{0.25em}
	\item \hd$\,{=}\,$\hdpercn{100} -- Initially introduced by Felix Hausdorff in 1914~\cite{Hausdorff1914-SetTheory} and popularized by Huttenlocher et al.\ in 1993~\cite{Huttenlocher1993-HD-Algorithm}, \hd\ represents one of the earliest metrics for evaluating the spatial distance between objects and quantifying the underlying dissimilarities. It is the only one that is mathematically well-defined even in the discretized space because it reports the maximal distance of the maximum of each directed distance, i.e.\ from $A$ to $B$ ($D_{\!AB}$) and from $B$ to $A$ ($D_{BA}$). The implementation is consistent across all open-source tools, except for \misevaltt, which does not account for the pixel/voxel size when computing the Euclidean distance transform but assumes an isotropic grid of unit size.
	\item \hdpercp\ -- This metric was first defined by Huttenlocher et al.~\cite{Huttenlocher1993-HD-Algorithm} as the maximum of both directed percentile distances. \metricsreloadedtt\ and \monaitt\ calculate it by sorting a set of distances $D_{\!AB}$ of cardinality $N$ in ascending order and selecting the value at position $(p/100){\cdot}N$ (rounded to the nearest integer), yielding $[D_{\!AB}]_p$, and vice versa $[D_{BA}]_p$. On the other hand, \googledeepmindtt\ and \pymiatt\ adopt a more general approach that accounts for the size of boundary elements (i.e.\ do not assume that the query points are uniformly distributed across the boundary). Other variations include the calculation of percentiles on the union of both directed sets of distances, which is implemented by \segmetricstt, \medpytt\ and \evaluatesegmentationtt. All these tools calculate the maximal value between the two directed percentile distances, while \plastimatchtt\ reports the average of the two instead. While \evaluatesegmentationtt\ incorporates optimization strategies for distance-based metrics~\cite{Taha2015-HD-Optimization}, these can only be applied to \hd\,{=}\,\hdpercn{100} and do not generalize to \hdpercp, therefore producing skewed, biased and non-deterministic results due to distorted distance distributions.
	\item \masd\ -- Although the origin of this metric traces back to Sluimer et al.~\cite{Sluimer2005-MASD}, its mathematical definition varies across implementations. Most tools -- \evaluatesegmentationtt, \metricsreloadedtt, \plastimatchtt, \pymiatt\ and \simpleitktt\ -- compute the mean by summing the distances in each directed set, then dividing by the number of query points in that set, and finally averaging the two directed results. In contrast, \animatt\ reports the maximum of the two directed averages, while \googledeepmindtt\ employs a more sophisticated strategy by computing a weighted average based on the boundary element size.
	\item \assd\ -- Two primary approaches exist to compute this metric that was originally defined by Lamecker et al.~\cite{Lamecker2004-ASSD}. \animatt, \medpytt, \metricsreloadedtt, \monaitt\ and \segmetricstt\ calculate the mean by summing both sets of directed distances and then dividing by the total number of distances across both sets. In contrast, \googledeepmindtt\ employs a weighted averaging strategy by assigning weights based on the boundary element size across both directed distance sets.
	\item \nsd\ -- Recently introduced by Nikolov et al.~\cite{Nikolov2021-DeepMind}, this metric differs from the four established metrics discussed above in that it is relative rather than absolute. \nsd, also referred to as the surface Dice, measures the proportion of overlap between the two segmentation boundaries, \segAboundary\ and \segBboundary, within a user-defined margin of error~$\tau$. A region $\msegAboundary^{(\pm\tau)}$ is first defined both inward~($-\tau$) and outward~($+\tau$) of boundary \segAboundary, and \nsd\ is then computed as the ratio of \segBboundary\ within $\msegAboundary^{(\pm\tau)}$, and vice versa, against the total size of both boundaries, represented by the contour circumference in 2D and surface area in 3D. Although the authors of \nsd\ explicitly highlighted the importance of accounting for the boundary element size in accurate \nsd\ computation~\cite{Nikolov2021-DeepMind}, implementations vary across tools. \metricsreloadedtt\ and \monaitt\ compute \nsd\ by counting the number of distances below $\tau$ and then dividing this count by the total number of distances. In contrast, \googledeepmindtt\ and \pymiatt\ perform weighting of the computation by the boundary element size, then summing the sizes of elements with distances below $\tau$, and finally dividing by the total size of all boundary elements (i.e.\ circumference in 2D and surface area in 3D).
	\item \biou\ -- This relative metric was recently introduced by Cheng et al.~\cite{Cheng2021BoundaryIoU}, and is currently supported only by \metricsreloadedtt. Its main feature is that it increases the sensitivity to boundary segmentation errors compared to conventional counting-based metrics such as IoU or \dsc, which often saturate when large segmentations exhibit substantial bulk overlap. A region $\msegAboundary^{(-\tau)}$ is first defined only inward~($-\tau$) of \segAboundary, and similarly a region $\msegBboundary^{(-\tau)}$ is defined for \segBboundary. Then, \biou\ is computed by measuring the intersection of $\msegAboundary^{(-\tau)}$ and $\msegBboundary^{(-\tau)}$ over their union. It can be therefore viewed as a hybrid between counting- and distance-based metrics, concentrating on the overlap of boundary regions rather than the whole segmentation mask. While the mathematical definitions remain consistent, \metricsreloadedtt\ performs calculations on a discretized grid, but with a flaw in its programming code: the pixel/voxel size is not properly accounted for.
\end{itemize}

\subsubsection{Edge case handling}
The results of probing the open-source tools with empty segmentations and the symmetry checks are summarized in Table~\ref{tab:edge-cases}. Our analysis revealed that most tools return \textit{not a number} (\texttt{NaN}), infinite values, large integers or error messages under these edge cases. However, we observed inconsistencies in edge case handling for \animatt\ (all metrics), \googledeepmindtt\ (\masd), \misevaltt\ (\hd), \plastimatchtt\ (\dsc, \hd\ and \masd), \pymiatt\ (\nsd), and \segmetricstt\ (\hd\ and \assd). Particularly misleading is the behaviour of \misevaltt, which returns random positive decimal values (\PosDec) when exactly one segmentation is empty and zero when both are empty. Additionally, all distance-based metrics analyzed in this study are symmetric, meaning the same value should be obtained when the input segmentations are swapped. We found that this is violated solely in \evaluatesegmentationtt\ for \hdpercn{95} and \masd\ due to non-deterministic calculations (Note~\ref{note:evaluatesegmentation} in Appendix).
\begin{table*}[!b]
	\caption{Edge case handling considering the input to open-source tools are two segmentations $A$ and $B$. O1--O3 denote the output of each tool when $A$ is an empty mask (O1), $B$ is empty mask (O2) or both $A$ and $B$ are empty masks (O3), while S denotes whether the resulting metric value remains unchanged when the inputs are swapped. Besides all distance-based metrics, \dsc\ is also included to provide a more comprehensive overview.~\textsuperscript{\ddag}Note that although \googledeepmindtt\ provides all necessary functions for its computation, it does not originally support \assd, and we therefore omit the analysis of \assd\ edge case handling for this tool.}
	\vspace{0.25cm}
	\label{tab:edge-cases}

\centering
\newcommand{\toolNameDOIX}[2]{\href{#2}{\texttt{#1}}}
\def\nameAnima{\toolNameDOIX{Anima}{https://doi.org/10.1038/s41598-018-31911-7}}
\def\nameEvaluateSegmentation{\toolNameDOIX{EvaluateSegmentation}{https://doi.org/10.1186/s12880-015-0068-x}}
\def\nameDeepMind{\toolNameDOIX{Google DeepMind}{https://doi.org/10.2196/26151}}
\def\nameMedPy{\toolNameDOIX{MedPy}{https://doi.org/10.5281/zenodo.2565940}}
\def\nameMetricsReloaded{\toolNameDOIX{Metrics Reloaded}{https://doi.org/10.1038/s41592-023-02151-z}}
\def\nameMISeval{\toolNameDOIX{MISeval}{https://doi.org/10.3233/shti220391}}
\def\nameMONAI{\toolNameDOIX{MONAI}{https://doi.org/10.48550/arXiv.2211.02701}}
\def\namePlastimatch{\toolNameDOIX{Plastimatch}{https://doi.org/10.1118/1.4961121}}
\def\namepymia{\toolNameDOIX{pymia}{https://doi.org/10.1016/j.cmpb.2020.105796}}
\def\nameSegMetrics{\toolNameDOIX{seg-metrics}{https://doi.org/10.1101/2024.02.22.24303215}}
\def\nameSimpleITK{\toolNameDOIX{SimpleITK}{https://doi.org/10.3389/fninf.2013.00045}}

\newcommand{\cellfmt}[4]{\footnotesize{#1} & \footnotesize{#2} & \footnotesize{#3} & \footnotesize{#4}}

\newcolumntype{C}[1]{>{\centering\let\newline\\\arraybackslash\hspace{0pt}}m{#1}}
\newcolumntype{Y}{>{\centering\arraybackslash}X}
\renewcommand{\arraystretch}{1.3} 
\small
\def\Off{\textsuperscript{~\,}} 
\newcommand{\sm}{\scalebox{0.75}[1.0]{\( - \)}}
\def\PosDec{\texttt{$\mathbb R_{>0}$}} 
\def\Nan{\texttt{NaN}} 
\def\Inf{\texttt{$\infty$}} 
\def\NegInf{\texttt{${\sm}\infty$}} 
\def\InfS{\texttt{\textsuperscript{\,}$\infty${\textsuperscript{\textasteriskcentered}}}} 
\def\Warn{\textsuperscript{\texttt{W}}} 
\def\Err{\texttt{E}} 
\def\ErrS{\texttt{\textsuperscript{~}E\textsuperscript{\textasteriskcentered}}} 
\setlength{\tabcolsep}{0pt}
\def\colM{2.1cm}
\def\colB{0.1cm}
%
\begin{tabularx}{\linewidth}{l C{\colB} *{4}{Y} C{\colB} *{4}{Y} C{\colB} *{4}{Y} C{\colB} *{4}{Y} C{\colB} *{4}{Y} C{\colB} *{4}{Y}}
\hline \noalign{\smallskip}
&&
\multicolumn{4}{C{\colM}}{\dsc} &&
\multicolumn{4}{C{\colM}}{\hd}	&&
\multicolumn{4}{C{\colM}}{\masd}&&
\multicolumn{4}{C{\colM}}{\assd}&&
\multicolumn{4}{C{\colM}}{\nsd}	&& \multicolumn{4}{C{\colM}}{\biou}\\
%
\cline{3-6} \cline{8-11}  \cline{13-16} \cline{18-21} \cline{23-26} \cline{28-31}
\noalign{\smallskip}
%
Open-source tool	 &&
\cellfmt{O1}{O2}{O3}{S} &&
\cellfmt{O1}{O2}{O3}{S} &&
\cellfmt{O1}{O2}{O3}{S} &&
\cellfmt{O1}{O2}{O3}{S} &&
\cellfmt{O1}{O2}{O3}{S} &&
\cellfmt{O1}{O2}{O3}{S} \\
\noalign{\smallskip} \hline \noalign{\smallskip}
%
\nameAnima~\cite{Commowick2018Anima}&&
\cellfmt{\ErrS}{0}{\ErrS}{\cmark}	&&
\cellfmt{\ErrS}{\Nan}{\ErrS}{\cmark}&&
\cellfmt{\ErrS}{\Nan}{\ErrS}{\cmark}&&
\cellfmt{\ErrS}{\Nan}{\ErrS}{\cmark}\\
%
\nameEvaluateSegmentation~\cite{Taha2015-EvaluateSegmentation} && \cellfmt{\Err}{\Err}{\Err}{\cmark}	&& \cellfmt{\Err}{\Err}{\Err}{\xmark}	&& \cellfmt{\Err}{\Err}{\Err}{\xmark}	\\
%
\nameDeepMind~\cite{Nikolov2021-DeepMind} &&
\cellfmt{0}{0}{\Nan}{\cmark}		&&
\cellfmt{\Inf}{\Inf}{\Inf}{\cmark}	&&
\cellfmt{\Nan}{\Nan}{\Nan}{\cmark}	&&
\multicolumn{4}{c}{\ddag}          &&
\cellfmt{0}{0}{\Off\Nan\Warn}{\cmark} 	\\
%
\nameMedPy~\cite{Maier2019MedPy}	&&
\cellfmt{0}{0}{1}{\cmark}			&&
\cellfmt{\Err}{\Err}{\Err}{\cmark}	&&
& & &								&&
\cellfmt{\Err}{\Err}{\Err}{\cmark}	\\
%
\nameMetricsReloaded~\cite{Maier-Hein2024-Metrics-Reloaded} &&
\cellfmt{0}{0}{\Off1\Warn}{\cmark}	&&
\cellfmt{\Off\Nan\Warn}{\Off\Nan\Warn}{\Off0\Warn}{\cmark}		&&
\cellfmt{\Off\Nan\Warn}{\Off\Nan\Warn}{\Off0\Warn}{\cmark}		&&
\cellfmt{\Off\Nan\Warn}{\Off\Nan\Warn}{\Off0\Warn}{\cmark}		&&
\cellfmt{0}{0}{\Off1\Warn}{\cmark} &&
\cellfmt{0}{0}{\Off1\Warn}{\cmark}		\\
%
\nameMISeval~\cite{Muller2022-MISeval} &&
\cellfmt{0}{0}{1}{\cmark}			&&
\cellfmt{\PosDec}{\PosDec}{0}{\cmark} \\
%
\nameMONAI~\cite{Cardoso2022-MONAI} &&
\cellfmt{0}{0}{1}{\cmark}			&&
\cellfmt{\Off\Nan\Warn}{\Off\Nan\Warn}{\Off\Nan\Warn}{\cmark} &&
& & &								&&
\cellfmt{\Off\Inf\Warn}{\Off\Inf\Warn}{\Off\Nan\Warn}{\cmark} &&
\cellfmt{\Off0\Warn}{\Off0\Warn}{\Off\Nan\Warn}{\cmark}		\\
%
\namePlastimatch~\cite{Zaffino2016-Plastimatch} &&
\cellfmt{0}{0}{\Off0\Warn}{\cmark}			&&
\cellfmt{\InfS}{\InfS}{\Err}{\cmark}&&
\cellfmt{\InfS}{\InfS}{\Err}{\cmark}\\
%
\namepymia~\cite{Jungo2021pymia} 	&&
\cellfmt{0}{0}{1}{\cmark}			&&
\cellfmt{\Off\Inf\Warn}{\Off\Inf\Warn}{\Off\Inf\Warn}{\cmark} && 
\cellfmt{\Off\Inf\Warn}{\Off\Inf\Warn}{\Off\Inf\Warn}{\cmark} &&
& & &								&&
\cellfmt{0}{0}{\Off\NegInf\Warn}{\cmark}\\
%
\nameSegMetrics~\cite{Jia2024-SegMetrics} &&
\cellfmt{\Off0\Warn}{\Off0\Warn}{\Off1\Warn}{\cmark}	&&
\cellfmt{0}{0}{0}{\cmark}&&
& & & 								&&
\cellfmt{0}{0}{0}{\cmark}\\
%
\nameSimpleITK~\cite{Lowekamp2013SimpleITK} &&
\cellfmt{0}{0}{\Inf}{\cmark}		&&
\cellfmt{\Err}{\Err}{\Err}{\cmark}	&&
\cellfmt{\Err}{\Err}{\Err}{\cmark}	\\
\noalign{\smallskip} \hline \noalign{\smallskip} 
\multicolumn{31}{p{0.985\textwidth}}{\footnotesize{%
    \Nan: not a number,
	\Err: error message, 
	\ErrS: empty string, no actual error message,
	\texttt{W}: additional warning message, 
	\InfS: a very large integer, \PosDec: a positive decimal number, \cmark: yes, \xmark: no}}\\
\noalign{\smallskip} \hline
\end{tabularx}
\end{table*}
\subsection{Empirical analysis}
\begin{table*}[!t]
	\caption{The differences ($\Delta$) in the distance-based metrics for the 2D dataset, as obtained by each open-source tool against the reference \googledeepmindtt\ implementation on 80 pairs of 2D segmentation masks, reported for two isotropic and one anisotropic pixel size as the range $\min{|}\max$ and mean $\pm$ standard deviation (SD).}
	\vspace{0.25cm}
	\label{tab:results-2D}
\vspace{-2pt}
\centering
\renewcommand{\arraystretch}{1.05}
\setlength{\tabcolsep}{0pt}
\newcommand{\colS}{\hspace*{0.3cm}}
\newcommand{\colT}{1.5cm}
\newcolumntype{x}[1]{>{\centering\arraybackslash\hspace{0pt}}p{#1}}
\newcolumntype{C}[1]{>{\centering\let\newline\\\arraybackslash\hspace{0pt}}m{#1}}
\def\clen{2.2cm}
\newcommand{\sm}{\scalebox{0.5}[1.0]{\( - \)}}
\newcommand{\fm}{~\,}

\def\minmax{~~Min\,$|$\,Max}
\def\meanstd{Mean$\,{\pm}\,$SD}
\begin{small}
\begin{tabularx}{\linewidth}{l r C{\clen}C{\clen} r C{\clen}C{\clen} r C{\clen}C{\clen}}
    \hline \noalign{\smallskip}
    & & \multicolumn{8}{c}{Pixel size} \\
    \noalign{\smallskip} \cline{3-10} \noalign{\smallskip}
    Open-source tool & &
    \multicolumn{2}{c}{{(1.0, 1.0) mm}} & &
    \multicolumn{2}{c}{{(0.07, 0.07) mm}} & &
    \multicolumn{2}{c}{{(0.07, 1.0) mm}} \\
    %
    %
\noalign{\smallskip} \hline \noalign{\bigskip}
$\bf{\Delta\,HD}$ (mm) & \colS & \multicolumn{1}{c}{\minmax} & \multicolumn{1}{c}{\meanstd} & \colS & \minmax & \meanstd & \colS & \minmax & \meanstd \\
\noalign{\smallskip} \hline \noalign{\smallskip}
\animatt                &  & ${\sm}1.29\,{|}\,\bf{0.00}$ & ${\sm}0.02\,{\pm}\,0.14$ &  & $\bf{0.00}\,{|}\,\bf{0.00}$ & $\bf{0.00}\,{\pm}\,\bf{0.00}$ &  & $\bf{0.00}\,{|}\,\bf{0.00}$ & $\bf{0.00}\,{\pm}\,\bf{0.00}$ \\	
\evaluatesegmentationtt &  & ${\sm}1.29\,{|}\,\bf{0.00}$ & ${\sm}0.02\,{\pm}\,0.14$ &  & $\bf{0.00}\,{|}\,\bf{0.00}$ & $\bf{0.00}\,{\pm}\,\bf{0.00}$ &  & $\bf{0.00}\,{|}\,\bf{0.00}$ & $\bf{0.00}\,{\pm}\,\bf{0.00}$ \\	
\medpytt                &  & $\bf{0.00}\,{|}\,\bf{0.00}$ & $\bf{0.00}\,{\pm}\,\bf{0.00}$ &  & $\bf{0.00}\,{|}\,\bf{0.00}$ & $\bf{0.00}\,{\pm}\,\bf{0.00}$ &  & $\bf{0.00}\,{|}\,\bf{0.00}$ & $\bf{0.00}\,{\pm}\,\bf{0.00}$ \\	
\metricsreloadedtt      &  & $\bf{0.00}\,{|}\,\bf{0.00}$ & $\bf{0.00}\,{\pm}\,\bf{0.00}$ &  & $\bf{0.00}\,{|}\,\bf{0.00}$ & $\bf{0.00}\,{\pm}\,\bf{0.00}$ &  & $\bf{0.00}\,{|}\,\bf{0.00}$ & $\bf{0.00}\,{\pm}\,\bf{0.00}$ \\	
\misevaltt              &  & ${\sm}1.29\,{|}\,\bf{0.00}$ & ${\sm}0.02\,{\pm}\,0.14$ &  & $44.6\,{|}\,2436$ & $1138\,{\pm}\,524$ &  & $0.99\,{|}\,1429$ & $439\,{\pm}\,333$ \\	
\monaitt                &  & $\bf{0.00}\,{|}\,\bf{0.00}$ & $\bf{0.00}\,{\pm}\,\bf{0.00}$ &  & $\bf{0.00}\,{|}\,\bf{0.00}$ & $\bf{0.00}\,{\pm}\,\bf{0.00}$ &  & $\bf{0.00}\,{|}\,\bf{0.00}$ & $\bf{0.00}\,{\pm}\,\bf{0.00}$ \\	
\plastimatchtt          &  & $\bf{0.00}\,{|}\,\bf{0.00}$ & $\bf{0.00}\,{\pm}\,\bf{0.00}$ &  & $\bf{0.00}\,{|}\,\bf{0.00}$ & $\bf{0.00}\,{\pm}\,\bf{0.00}$ &  & $\bf{0.00}\,{|}\,\bf{0.00}$ & $\bf{0.00}\,{\pm}\,\bf{0.00}$ \\	
\pymiatt                &  & ${\sm}1.29\,{|}\,\bf{0.00}$ & ${\sm}0.02\,{\pm}\,0.14$ &  & $\bf{0.00}\,{|}\,\bf{0.00}$ & $\bf{0.00}\,{\pm}\,\bf{0.00}$ &  & $\bf{0.00}\,{|}\,\bf{0.00}$ & $\bf{0.00}\,{\pm}\,\bf{0.00}$ \\	
\segmetricstt           &  & ${\sm}0.52\,{|}\,\bf{0.00}$ & ${\sm}0.01\,{\pm}\,0.06$ &  & $\bf{0.00}\,{|}\,\bf{0.00}$ & $\bf{0.00}\,{\pm}\,\bf{0.00}$ &  & $\bf{0.00}\,{|}\,\bf{0.00}$ & $\bf{0.00}\,{\pm}\,\bf{0.00}$ \\	
\simpleitktt            &  & ${\sm}1.29\,{|}\,\bf{0.00}$ & ${\sm}0.02\,{\pm}\,0.14$ &  & $\bf{0.00}\,{|}\,\bf{0.00}$ & $\bf{0.00}\,{\pm}\,\bf{0.00}$ &  & $\bf{0.00}\,{|}\,\bf{0.00}$ & $\bf{0.00}\,{\pm}\,\bf{0.00}$ \\	
\noalign{\smallskip} \hline \noalign{\bigskip}
$\bf{\Delta\,HD_{95}}$ (mm) & \colS & \multicolumn{1}{c}{\minmax} & \multicolumn{1}{c}{\meanstd} & \colS & \minmax & \meanstd & \colS & \minmax & \meanstd \\
\noalign{\smallskip} \hline \noalign{\smallskip}
\evaluatesegmentationtt &  & ${\sm}\bf{73.4}\,{|}\,27.1$ & ${\sm}\bf{1.45}\,{\pm}\,\bf{14.4}$ &  & ${\sm}66.9\,{|}\,28.2$ & ${\sm}0.37\,{\pm}\,12.5$ &  & ${\sm}\bf{67.3}\,{|}\,34.0$ & ${\sm}\bf{0.09}\,{\pm}\,\bf{13.0}$ \\	
\medpytt                &  & ${\sm}116\,{|}\,0.55$ & ${\sm}8.59\,{\pm}\,19.5$ &  & ${\sm}38.6\,{|}\,{\sm}0.06$ & ${\sm}3.09\,{\pm}\,6.99$ &  & ${\sm}117\,{|}\,0.85$ & ${\sm}8.74\,{\pm}\,20.4$ \\	
\metricsreloadedtt      &  & ${\sm}84.8\,{|}\,1.50$ & ${\sm}2.83\,{\pm}\,12.0$ &  & ${\sm}\bf{1.59}\,{|}\,0.58$ & $\bf{0.01}\,{\pm}\,\bf{0.22}$ &  & ${\sm}75.7\,{|}\,1.14$ & ${\sm}3.34\,{\pm}\,11.8$ \\	
\monaitt                &  & ${\sm}84.8\,{|}\,1.50$ & ${\sm}2.83\,{\pm}\,12.0$ &  & ${\sm}\bf{1.59}\,{|}\,0.58$ & $\bf{0.01}\,{\pm}\,\bf{0.22}$ &  & ${\sm}75.7\,{|}\,1.14$ & ${\sm}3.34\,{\pm}\,11.8$ \\	
\plastimatchtt          &  & ${\sm}106\,{|}\,\bf{0.00}$ & ${\sm}36.7\,{\pm}\,22.8$ &  & ${\sm}90.7\,{|}\,{\sm}0.13$ & ${\sm}36.0\,{\pm}\,20.6$ &  & ${\sm}100.0\,{|}\,\bf{0.00}$ & ${\sm}37.4\,{\pm}\,22.4$ \\	
\pymiatt                &  & ${\sm}118\,{|}\,{\sm}0.33$ & ${\sm}14.4\,{\pm}\,23.3$ &  & ${\sm}117\,{|}\,{\sm}0.21$ & ${\sm}13.1\,{\pm}\,21.7$ &  & ${\sm}117\,{|}\,{\sm}0.30$ & ${\sm}13.0\,{\pm}\,21.5$ \\	
\segmetricstt           &  & ${\sm}117\,{|}\,0.30$ & ${\sm}10.4\,{\pm}\,20.7$ &  & ${\sm}75.4\,{|}\,{\sm}\bf{0.02}$ & ${\sm}4.81\,{\pm}\,11.9$ &  & ${\sm}117\,{|}\,0.80$ & ${\sm}8.77\,{\pm}\,20.4$ \\	
\noalign{\smallskip} \hline \noalign{\bigskip}
$\bf{\Delta\,MASD}$ (mm) & \colS & \multicolumn{1}{c}{\minmax} & \multicolumn{1}{c}{\meanstd} & \colS & \minmax & \meanstd & \colS & \minmax & \meanstd \\
\noalign{\smallskip} \hline \noalign{\smallskip}
\animatt                &  & $\bf{0.01}\,{|}\,64.8$ & $12.1\,{\pm}\,12.0$ &  & $\bf{0.03}\,{|}\,60.2$ & $14.9\,{\pm}\,12.7$ &  & ${\sm}\bf{2.27}\,{|}\,48.6$ & $12.0\,{\pm}\,11.2$ \\	
\evaluatesegmentationtt &  & ${\sm}18.4\,{|}\,13.7$ & ${\sm}5.19\,{\pm}\,4.85$ &  & ${\sm}54.2\,{|}\,11.9$ & ${\sm}8.25\,{\pm}\,10.6$ &  & ${\sm}20.2\,{|}\,7.65$ & ${\sm}5.48\,{\pm}\,5.73$ \\	
\metricsreloadedtt      &  & ${\sm}7.63\,{|}\,\bf{3.78}$ & ${\sm}\bf{0.77}\,{\pm}\,\bf{1.36}$ &  & ${\sm}0.43\,{|}\,\bf{1.88}$ & $\bf{0.06}\,{\pm}\,\bf{0.35}$ &  & ${\sm}8.09\,{|}\,\bf{4.84}$ & ${\sm}\bf{1.13}\,{\pm}\,\bf{2.05}$ \\	
\plastimatchtt          &  & ${\sm}7.63\,{|}\,\bf{3.78}$ & ${\sm}\bf{0.77}\,{\pm}\,\bf{1.36}$ &  & ${\sm}0.43\,{|}\,\bf{1.88}$ & $\bf{0.06}\,{\pm}\,\bf{0.35}$ &  & ${\sm}8.09\,{|}\,\bf{4.84}$ & ${\sm}\bf{1.13}\,{\pm}\,\bf{2.05}$ \\	
\pymiatt                &  & ${\sm}18.4\,{|}\,13.7$ & ${\sm}5.19\,{\pm}\,4.85$ &  & ${\sm}35.5\,{|}\,14.2$ & ${\sm}6.48\,{\pm}\,6.50$ &  & ${\sm}20.0\,{|}\,16.4$ & ${\sm}5.76\,{\pm}\,5.35$ \\	
\simpleitktt            &  & ${\sm}18.4\,{|}\,13.7$ & ${\sm}5.19\,{\pm}\,4.85$ &  & ${\sm}35.5\,{|}\,14.2$ & ${\sm}6.48\,{\pm}\,6.50$ &  & ${\sm}20.0\,{|}\,16.4$ & ${\sm}5.76\,{\pm}\,5.35$ \\	
\noalign{\smallskip} \hline \noalign{\bigskip}
$\bf{\Delta\,ASSD}$ (mm) & \colS & \multicolumn{1}{c}{\minmax} & \multicolumn{1}{c}{\meanstd} & \colS & \minmax & \meanstd & \colS & \minmax & \meanstd \\
\noalign{\smallskip} \hline \noalign{\smallskip}
\animatt                &  & ${\sm}51.4\,{|}\,6.09$ & ${\sm}11.4\,{\pm}\,14.4$ &  & ${\sm}52.8\,{|}\,2.38$ & ${\sm}11.2\,{\pm}\,15.2$ &  & ${\sm}52.1\,{|}\,\bf{7.22}$ & ${\sm}12.2\,{\pm}\,15.2$ \\	
\medpytt                &  & ${\sm}\bf{12.5}\,{|}\,7.63$ & ${\sm}\bf{1.34}\,{\pm}\,\bf{2.21}$ &  & ${\sm}\bf{0.88}\,{|}\,4.26$ & $\bf{0.09}\,{\pm}\,\bf{0.71}$ &  & ${\sm}15.2\,{|}\,10.4$ & ${\sm}\bf{2.09}\,{\pm}\,\bf{3.75}$ \\	
\metricsreloadedtt      &  & ${\sm}\bf{12.5}\,{|}\,7.63$ & ${\sm}\bf{1.34}\,{\pm}\,\bf{2.21}$ &  & ${\sm}\bf{0.88}\,{|}\,4.26$ & $\bf{0.09}\,{\pm}\,\bf{0.71}$ &  & ${\sm}15.2\,{|}\,10.4$ & ${\sm}\bf{2.09}\,{\pm}\,\bf{3.75}$ \\	
\monaitt                &  & ${\sm}\bf{12.5}\,{|}\,7.63$ & ${\sm}\bf{1.34}\,{\pm}\,\bf{2.21}$ &  & ${\sm}\bf{0.88}\,{|}\,4.26$ & $\bf{0.09}\,{\pm}\,\bf{0.71}$ &  & ${\sm}15.2\,{|}\,10.4$ & ${\sm}\bf{2.09}\,{\pm}\,\bf{3.75}$ \\	
\segmetricstt           &  & ${\sm}17.1\,{|}\,\bf{1.89}$ & ${\sm}2.57\,{\pm}\,3.00$ &  & ${\sm}9.04\,{|}\,\bf{0.62}$ & ${\sm}0.46\,{\pm}\,1.27$ &  & ${\sm}\bf{14.2}\,{|}\,8.53$ & ${\sm}2.17\,{\pm}\,3.50$ \\	
\noalign{\smallskip} \hline \noalign{\bigskip}
$\bf{\Delta\,NSD_{\,\tau=2\,mm}}$ (\%pt) & \colS & \multicolumn{1}{c}{\minmax} & \multicolumn{1}{c}{\meanstd} & \colS & \minmax & \meanstd & \colS & \minmax & \meanstd \\
\noalign{\smallskip} \hline \noalign{\smallskip}
\metricsreloadedtt      &  & ${\sm}\bf{8.36}\,{|}\,\bf{11.1}$ & $\bf{1.10}\,{\pm}\,\bf{2.88}$ &  & ${\sm}\bf{2.13}\,{|}\,\bf{1.35}$ & ${\sm}\bf{0.06}\,{\pm}\,\bf{0.57}$ &  & ${\sm}\bf{4.58}\,{|}\,\bf{15.4}$ & $\bf{2.82}\,{\pm}\,\bf{3.89}$ \\	
\monaitt                &  & ${\sm}\bf{8.36}\,{|}\,\bf{11.1}$ & $\bf{1.10}\,{\pm}\,\bf{2.88}$ &  & ${\sm}\bf{2.13}\,{|}\,\bf{1.35}$ & ${\sm}\bf{0.06}\,{\pm}\,\bf{0.57}$ &  & ${\sm}\bf{4.58}\,{|}\,\bf{15.4}$ & $\bf{2.82}\,{\pm}\,\bf{3.89}$ \\	
\pymiatt                &  & ${\sm}10.2\,{|}\,45.9$ & $18.4\,{\pm}\,11.4$ &  & ${\sm}8.14\,{|}\,51.6$ & $19.2\,{\pm}\,11.8$ &  & ${\sm}10.9\,{|}\,49.2$ & $18.9\,{\pm}\,11.7$ \\	
\noalign{\smallskip} \hline \noalign{\bigskip}
$\bf{\Delta\,BIoU_{\,\tau=2\,mm}}$ (\%pt) & \colS & \multicolumn{1}{c}{\minmax} & \multicolumn{1}{c}{\meanstd} & \colS & \minmax & \meanstd & \colS & \minmax & \meanstd \\
\noalign{\smallskip} \hline \noalign{\smallskip}
\metricsreloadedtt      &  & ${\sm}\bf{5.78}\,{|}\,\bf{6.09}$ & $\bf{1.91}\,{\pm}\,\bf{2.27}$ &  & ${\sm}\bf{64.7}\,{|}\,\bf{0.00}$ & ${\sm}\bf{29.0}\,{\pm}\,\bf{16.1}$ &  & ${\sm}\bf{18.7}\,{|}\,\bf{8.31}$ & ${\sm}\bf{7.41}\,{\pm}\,\bf{5.16}$ \\	
\noalign{\smallskip} \hline
\end{tabularx}
\end{small}
\end{table*}
\begin{table*}[!t]
	\caption{The differences ($\Delta$) in the distance-based metrics for the 3D dataset, as obtained by each open-source tool against the reference \googledeepmindtt\ implementation on 1,559 pairs of 3D segmentation masks, reported for two isotropic and one anisotropic voxel size as the range $\min{|}\max$ and mean $\pm$ standard deviation (SD).}
	\vspace{0.25cm}
	\label{tab:results-3D}
\vspace{-2pt}
\centering
\renewcommand{\arraystretch}{1.05}
\setlength{\tabcolsep}{0pt}
\newcommand{\colS}{\hspace*{0.3cm}}
\newcommand{\colT}{1.5cm}
\newcolumntype{x}[1]{>{\centering\arraybackslash\hspace{0pt}}p{#1}}
\newcolumntype{C}[1]{>{\centering\let\newline\\\arraybackslash\hspace{0pt}}m{#1}}
\def\clen{2.2cm}
\newcommand{\sm}{\scalebox{0.5}[1.0]{\( - \)}}
\newcommand{\fm}{~\,}

\def\minmax{~~Min\,$|$\,Max}
\def\meanstd{Mean$\,{\pm}\,$SD}
\begin{small}
\begin{tabularx}{\linewidth}{l r C{\clen}C{\clen} r C{\clen}C{\clen} r C{\clen}C{\clen}}
    \hline \noalign{\smallskip}
    & & \multicolumn{8}{c}{Voxel size} \\
    \noalign{\smallskip} \cline{3-10} \noalign{\smallskip}
    Open-source tool & &
    \multicolumn{2}{c}{{(1.0, 1.0, 1.0) mm}} & &
    \multicolumn{2}{c}{{(2.0, 2.0, 2.0) mm}} & &
    \multicolumn{2}{c}{{(0.5, 0.5, 2.0) mm}} \\
    %
    %
\noalign{\smallskip} \hline \noalign{\bigskip}
$\bf{\Delta\,HD}$ (mm) & \colS & \multicolumn{1}{c}{\minmax} & \multicolumn{1}{c}{\meanstd} & \colS & \minmax & \meanstd & \colS & \minmax & \meanstd \\
\noalign{\smallskip} \hline \noalign{\smallskip}
\animatt                &  & ${\sm}6.16\,{|}\,0.65$ & ${\sm}0.01\,{\pm}\,0.24$ &  & ${\sm}5.62\,{|}\,1.73$ & $0.01\,{\pm}\,0.27$ &  & ${\sm}6.35\,{|}\,1.85$ & ${\sm}0.01\,{\pm}\,0.27$ \\	
\evaluatesegmentationtt &  & ${\sm}6.16\,{|}\,0.65$ & ${\sm}0.01\,{\pm}\,0.24$ &  & ${\sm}5.62\,{|}\,1.73$ & $0.01\,{\pm}\,0.27$ &  & ${\sm}6.35\,{|}\,1.85$ & ${\sm}0.01\,{\pm}\,0.27$ \\	
\medpytt                &  & ${\sm}0.52\,{|}\,0.65$ & $0.01\,{\pm}\,0.06$ &  & ${\sm}0.72\,{|}\,1.73$ & $0.03\,{\pm}\,0.15$ &  & ${\sm}0.39\,{|}\,1.85$ & $0.02\,{\pm}\,0.09$ \\	
\metricsreloadedtt      &  & ${\sm}0.52\,{|}\,0.65$ & $0.01\,{\pm}\,0.06$ &  & ${\sm}0.72\,{|}\,1.73$ & $0.03\,{\pm}\,0.15$ &  & ${\sm}0.39\,{|}\,1.85$ & $0.02\,{\pm}\,0.09$ \\	
\misevaltt              &  & ${\sm}6.16\,{|}\,0.65$ & ${\sm}0.01\,{\pm}\,0.24$ &  & ${\sm}124\,{|}\,{\sm}0.59$ & ${\sm}5.09\,{\pm}\,6.63$ &  & ${\sm}123\,{|}\,43.0$ & $2.87\,{\pm}\,7.06$ \\	
\monaitt                &  & ${\sm}0.52\,{|}\,0.65$ & $0.01\,{\pm}\,0.06$ &  & ${\sm}0.72\,{|}\,1.73$ & $0.03\,{\pm}\,0.15$ &  & ${\sm}0.39\,{|}\,1.85$ & $0.02\,{\pm}\,0.09$ \\	
\plastimatchtt          &  & ${\sm}0.52\,{|}\,0.65$ & $0.01\,{\pm}\,0.06$ &  & ${\sm}0.72\,{|}\,1.73$ & $0.03\,{\pm}\,0.16$ &  & ${\sm}0.36\,{|}\,1.85$ & $0.02\,{\pm}\,0.09$ \\	
\pymiatt                &  & $\bf{0.00}\,{|}\,\bf{0.00}$ & $\bf{0.00}\,{\pm}\,\bf{0.00}$ &  & $\bf{0.00}\,{|}\,\bf{0.00}$ & $\bf{0.00}\,{\pm}\,\bf{0.00}$ &  & $\bf{0.00}\,{|}\,\bf{0.00}$ & $\bf{0.00}\,{\pm}\,\bf{0.00}$ \\	
\segmetricstt           &  & ${\sm}0.73\,{|}\,9.57$ & $0.04\,{\pm}\,0.35$ &  & ${\sm}0.76\,{|}\,3.53$ & $0.03\,{\pm}\,0.18$ &  & ${\sm}0.60\,{|}\,2.24$ & $0.02\,{\pm}\,0.11$ \\	
\simpleitktt            &  & ${\sm}6.16\,{|}\,0.65$ & ${\sm}0.01\,{\pm}\,0.24$ &  & ${\sm}5.62\,{|}\,1.73$ & $0.01\,{\pm}\,0.27$ &  & ${\sm}6.35\,{|}\,1.85$ & ${\sm}0.01\,{\pm}\,0.27$ \\	
\noalign{\smallskip} \hline \noalign{\bigskip}
$\bf{\Delta\,HD_{95}}$ (mm) & \colS & \multicolumn{1}{c}{\minmax} & \multicolumn{1}{c}{\meanstd} & \colS & \minmax & \meanstd & \colS & \minmax & \meanstd \\
\noalign{\smallskip} \hline \noalign{\smallskip}
\evaluatesegmentationtt &  & ${\sm}0.13\,{|}\,17.1$ & $2.96\,{\pm}\,2.75$ &  & ${\sm}0.25\,{|}\,16.5$ & $2.86\,{\pm}\,2.84$ &  & $\bf{0.00}\,{|}\,16.4$ & $2.97\,{\pm}\,2.74$ \\	
\medpytt                &  & ${\sm}17.3\,{|}\,1.00$ & ${\sm}0.67\,{\pm}\,1.61$ &  & ${\sm}17.0\,{|}\,2.00$ & ${\sm}0.50\,{\pm}\,1.66$ &  & ${\sm}16.2\,{|}\,2.00$ & ${\sm}0.43\,{\pm}\,1.30$ \\	
\metricsreloadedtt      &  & ${\sm}4.42\,{|}\,1.88$ & $0.22\,{\pm}\,0.28$ &  & ${\sm}6.48\,{|}\,2.52$ & $0.35\,{\pm}\,0.52$ &  & ${\sm}4.52\,{|}\,8.00$ & $0.42\,{\pm}\,0.84$ \\	
\monaitt                &  & ${\sm}4.42\,{|}\,1.88$ & $0.22\,{\pm}\,0.28$ &  & ${\sm}6.48\,{|}\,2.52$ & $0.35\,{\pm}\,0.52$ &  & ${\sm}4.52\,{|}\,8.00$ & $0.42\,{\pm}\,0.84$ \\	
\plastimatchtt          &  & ${\sm}115\,{|}\,1.00$ & ${\sm}1.80\,{\pm}\,5.74$ &  & ${\sm}114\,{|}\,2.00$ & ${\sm}1.65\,{\pm}\,5.71$ &  & ${\sm}111\,{|}\,2.20$ & ${\sm}1.67\,{\pm}\,5.55$ \\	
\pymiatt                &  & $\bf{0.00}\,{|}\,\bf{0.00}$ & $\bf{0.00}\,{\pm}\,\bf{0.00}$ &  & $\bf{0.00}\,{|}\,\bf{0.00}$ & $\bf{0.00}\,{\pm}\,\bf{0.00}$ &  & $\bf{0.00}\,{|}\,\bf{0.00}$ & $\bf{0.00}\,{\pm}\,\bf{0.00}$ \\	
\segmetricstt           &  & ${\sm}19.2\,{|}\,0.79$ & ${\sm}1.06\,{\pm}\,1.86$ &  & ${\sm}17.0\,{|}\,2.00$ & ${\sm}0.78\,{\pm}\,1.82$ &  & ${\sm}16.9\,{|}\,1.20$ & ${\sm}0.73\,{\pm}\,1.54$ \\	
\noalign{\smallskip} \hline \noalign{\bigskip}
$\bf{\Delta\,MASD}$ (mm) & \colS & \multicolumn{1}{c}{\minmax} & \multicolumn{1}{c}{\meanstd} & \colS & \minmax & \meanstd & \colS & \minmax & \meanstd \\
\noalign{\smallskip} \hline \noalign{\smallskip}
\animatt                &  & ${\sm}0.36\,{|}\,38.0$ & $0.74\,{\pm}\,1.70$ &  & $\bf{0.08}\,{|}\,40.8$ & $1.00\,{\pm}\,1.77$ &  & ${\sm}\bf{0.08}\,{|}\,32.6$ & $0.86\,{\pm}\,1.75$ \\	
\evaluatesegmentationtt &  & ${\sm}2.68\,{|}\,2.58$ & ${\sm}0.42\,{\pm}\,0.43$ &  & ${\sm}2.15\,{|}\,2.60$ & ${\sm}\bf{0.11}\,{\pm}\,\bf{0.49}$ &  & ${\sm}2.74\,{|}\,4.37$ & ${\sm}0.37\,{\pm}\,0.47$ \\	
\metricsreloadedtt      &  & ${\sm}\bf{0.24}\,{|}\,\bf{0.96}$ & $\bf{0.22}\,{\pm}\,\bf{0.09}$ &  & ${\sm}0.98\,{|}\,\bf{1.81}$ & $0.41\,{\pm}\,0.19$ &  & ${\sm}2.23\,{|}\,\bf{1.89}$ & $\bf{0.24}\,{\pm}\,\bf{0.27}$ \\	
\plastimatchtt          &  & ${\sm}\bf{0.24}\,{|}\,\bf{0.96}$ & $\bf{0.22}\,{\pm}\,\bf{0.09}$ &  & ${\sm}0.98\,{|}\,\bf{1.81}$ & $0.41\,{\pm}\,0.19$ &  & ${\sm}2.23\,{|}\,\bf{1.89}$ & $\bf{0.24}\,{\pm}\,\bf{0.27}$ \\	
\pymiatt                &  & ${\sm}2.68\,{|}\,2.58$ & ${\sm}0.42\,{\pm}\,0.43$ &  & ${\sm}2.15\,{|}\,2.60$ & ${\sm}\bf{0.11}\,{\pm}\,\bf{0.49}$ &  & ${\sm}2.74\,{|}\,3.01$ & ${\sm}0.40\,{\pm}\,0.44$ \\	
\simpleitktt            &  & ${\sm}2.68\,{|}\,2.58$ & ${\sm}0.42\,{\pm}\,0.43$ &  & ${\sm}2.15\,{|}\,2.60$ & ${\sm}\bf{0.11}\,{\pm}\,\bf{0.49}$ &  & ${\sm}2.74\,{|}\,3.01$ & ${\sm}0.40\,{\pm}\,0.44$ \\
\noalign{\smallskip} \hline \noalign{\bigskip}
$\bf{\Delta\,ASSD}$ (mm) & \colS & \multicolumn{1}{c}{\minmax} & \multicolumn{1}{c}{\meanstd} & \colS & \minmax & \meanstd & \colS & \minmax & \meanstd \\
\noalign{\smallskip} \hline \noalign{\smallskip}
\animatt                &  & ${\sm}22.8\,{|}\,\bf{0.65}$ & ${\sm}1.17\,{\pm}\,1.27$ &  & ${\sm}22.6\,{|}\,2.16$ & ${\sm}0.98\,{\pm}\,1.22$ &  & ${\sm}22.7\,{|}\,\bf{1.74}$ & ${\sm}1.07\,{\pm}\,1.26$ \\	
\medpytt                &  & ${\sm}\bf{0.49}\,{|}\,1.72$ & $0.23\,{\pm}\,0.10$ &  & ${\sm}\bf{1.61}\,{|}\,3.09$ & $0.43\,{\pm}\,0.22$ &  & ${\sm}4.43\,{|}\,2.97$ & $0.25\,{\pm}\,0.33$ \\	
\metricsreloadedtt      &  & ${\sm}\bf{0.49}\,{|}\,1.72$ & $0.23\,{\pm}\,0.10$ &  & ${\sm}\bf{1.61}\,{|}\,3.09$ & $0.43\,{\pm}\,0.22$ &  & ${\sm}4.43\,{|}\,2.97$ & $0.25\,{\pm}\,0.33$ \\	
\monaitt                &  & ${\sm}\bf{0.49}\,{|}\,1.72$ & $0.23\,{\pm}\,0.10$ &  & ${\sm}\bf{1.61}\,{|}\,3.09$ & $0.43\,{\pm}\,0.22$ &  & ${\sm}4.43\,{|}\,2.97$ & $0.25\,{\pm}\,0.33$ \\	
\segmetricstt           &  & ${\sm}1.80\,{|}\,0.86$ & $\bf{0.01}\,{\pm}\,\bf{0.16}$ &  & ${\sm}2.93\,{|}\,\bf{1.61}$ & $\bf{0.17}\,{\pm}\,\bf{0.29}$ &  & ${\sm}\bf{3.97}\,{|}\,1.95$ & $\bf{0.08}\,{\pm}\,\bf{0.25}$ \\
\noalign{\smallskip} \hline \noalign{\bigskip}
$\bf{\Delta\,NSD_{\,\tau=2\,mm}}$ (\%pt) & \colS & \multicolumn{1}{c}{\minmax} & \multicolumn{1}{c}{\meanstd} & \colS & \minmax & \meanstd & \colS & \minmax & \meanstd \\
\noalign{\smallskip} \hline \noalign{\smallskip}
\metricsreloadedtt      &  & ${\sm}16.6\,{|}\,2.56$ & ${\sm}2.71\,{\pm}\,2.12$ &  & ${\sm}37.5\,{|}\,3.77$ & ${\sm}4.62\,{\pm}\,4.38$ &  & ${\sm}19.2\,{|}\,7.00$ & ${\sm}3.52\,{\pm}\,3.59$ \\	
\monaitt                &  & ${\sm}16.6\,{|}\,2.56$ & ${\sm}2.71\,{\pm}\,2.12$ &  & ${\sm}37.5\,{|}\,3.77$ & ${\sm}4.62\,{\pm}\,4.38$ &  & ${\sm}19.2\,{|}\,7.00$ & ${\sm}3.52\,{\pm}\,3.59$ \\	
\pymiatt                &  & $\bf{0.00}\,{|}\,\bf{0.00}$ & $\bf{0.00}\,{\pm}\,\bf{0.00}$ &  & $\bf{0.00}\,{|}\,\bf{0.00}$ & $\bf{0.00}\,{\pm}\,\bf{0.00}$ &  & $\bf{0.00}\,{|}\,\bf{0.00}$ & $\bf{0.00}\,{\pm}\,\bf{0.00}$ \\
\noalign{\smallskip} \hline \noalign{\bigskip}
$\bf{\Delta\,BIoU_{\,\tau=2\,mm}}$ (\%pt) & \colS & \multicolumn{1}{c}{\minmax} & \multicolumn{1}{c}{\meanstd} & \colS & \minmax & \meanstd & \colS & \minmax & \meanstd \\
\noalign{\smallskip} \hline \noalign{\smallskip}
\metricsreloadedtt      &  & ${\sm}\bf{4.49}\,{|}\,\bf{8.24}$ & $\bf{1.12}\,{\pm}\,\bf{1.42}$ &  & ${\sm}\bf{8.32}\,{|}\,\bf{38.5}$ & $\bf{15.3}\,{\pm}\,\bf{9.60}$ &  & ${\sm}\bf{15.9}\,{|}\,\bf{22.2}$ & ${\sm}\bf{0.56}\,{\pm}\,\bf{5.15}$ \\
\noalign{\smallskip} \hline
\end{tabularx}
\end{small}
\end{table*}
\subsubsection{Metric value variability}
Directly comparing the raw metric values across implementations for each segmentation pair is impractical due to substantial disparities in the results. Instead, we analyzed the deviations of open-source tools by taking \googledeepmindtt\ as the reference implementation: $\Delta_{j,i} = m_{j,i} - m_{\textnormal{GDM},i}$, where $m_{j,i}$ is the metric value of $j$-th open-source tool and $m_{\textnormal{GDM},i}$ is the reference value of \googledeepmindtt\ when comparing values for $i$-th pair of segmentations. We selected \googledeepmindtt\ as the reference because it exhibited no major mathematical or boundary extraction pitfalls in our conceptual analysis, and provides implementations for nearly all metrics. Although \assd\ and \biou\ are originally not included, we implemented them using the functions available within the \googledeepmindtt\ package.
\par
The distance-based metrics can be divided into (1)~\textit{absolute metrics} (\hd, \hdpercp, \masd, \assd), measured in metric units (i.e.\ in our case in millimeters, mm), without an upper bound, with a lower value reflecting a greater similarity between two segmentations, and (2)~\textit{relative metrics} (\nsd, \biou), bounded between 0 and 1, unitless, with a higher value reflecting a greater similarity between two segmentations. In the context of this analysis, positive $\Delta_{j,i}$ values indicate an \textit{overestimation} of the metric value (i.e.\ over-pessimistic for absolute and over-optimistic for relative metrics), while negative $\Delta_{j,i}$ values indicate an \textit{underestimation} of the metric value (i.e.\ over-optimistic for absolute and over-pessimistic for relative metrics). Although both are undesirable, over-optimistic estimates are particularly concerning, as they may lead to incorrect conclusions, especially when compared to metric values reported in existing studies.
\par
The quantitative results for the 2D and 3D datasets, stratified by the pixel/voxel size, are summarized in Tables~\ref{tab:results-2D} and \ref{tab:results-3D}. We report the minimum, maximum, mean and standard deviation of metric value deviations to estimate their aggregated impact on realistic clinical segmentations. The analysis revealed notable differences across all metrics and both datasets, except for \hd, where deviations are relatively small for most tools. For non-unit pixel/voxel sizes, \misevaltt\ exhibits the largest deviations due to ignoring the image element size. For the remaining tools, deviations generally increase with non-unit image element sizes.
\par
A more detailed analysis of the results is provided in Appendix. Respectively for the 2D and 3D dataset, Fig.~\ref{fig:boxplots-2D} and \ref{fig:boxplots-3D} graphically illustrate the results from Tables~\ref{tab:results-2D} and \ref{tab:results-3D}, while Fig.~\ref{fig:stat-testing-2D} and Fig.~\ref{fig:stat-testing-3D} show statistical comparisons of deviation distributions using a two-sided paired Wilcoxon signed-rank test with the Bonferroni correction. They revealed significant differences between most tool pairs, except for \hd, which exhibits a lower variability due to its consistent mathematical definition and reduced sensitivity to boundary extraction, as it relies on a maximum-value statistics. Additionally, Fig.~\ref{fig:organs-groups} and Table~\ref{tab:results-per-organ-group} present the results for the 3D dataset, stratified by four organ categories: small, midsize, large and tubular.
\subsubsection{Computational efficiency}
\begin{figure*}[!ht]
	\centering
	\includegraphics[width=0.98\textwidth]{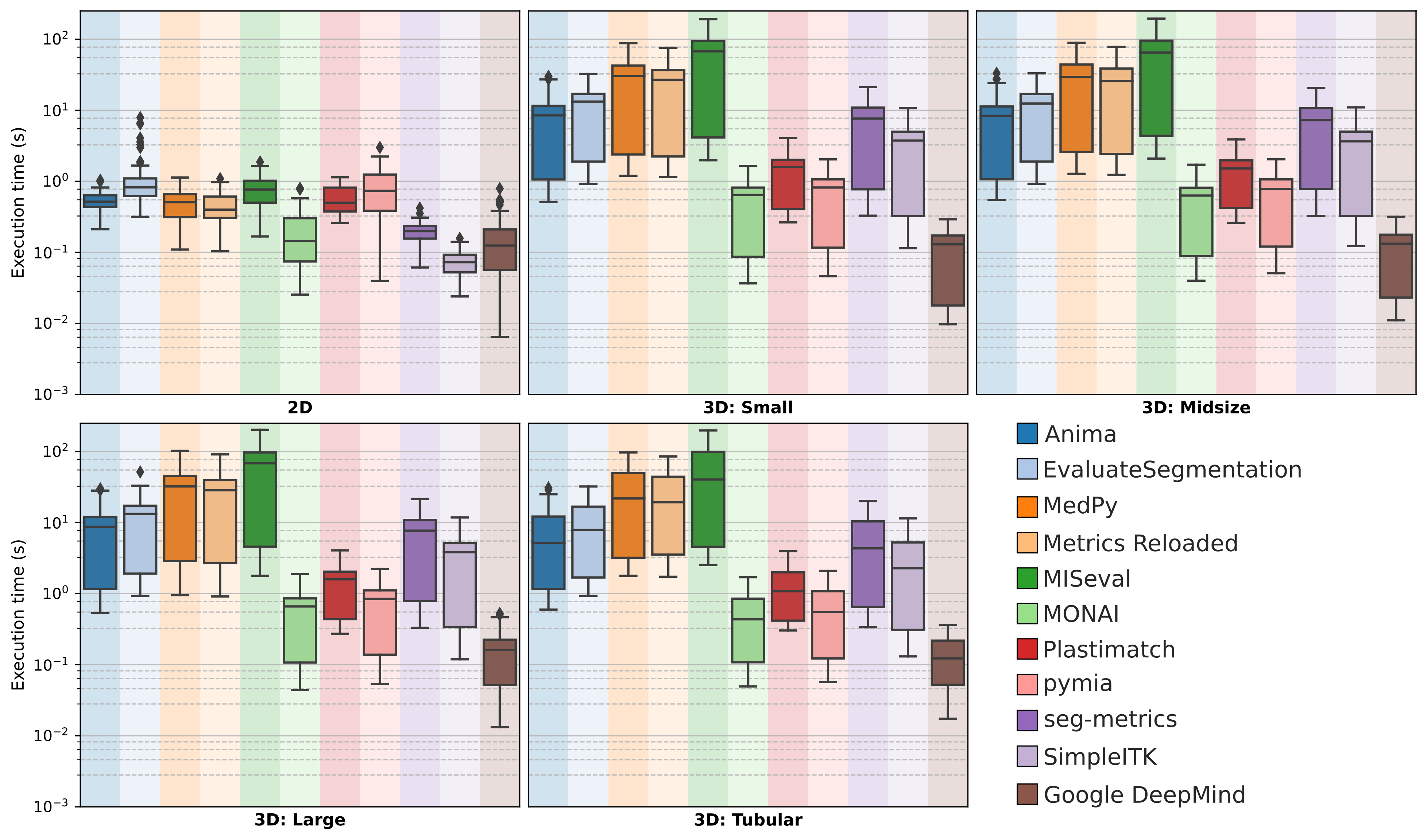}
	\vspace{0.25cm}
	\caption{Comparison of the \hd\ computational efficiency across all 11 open-source tools on 2D and 3D datasets, reported for the pixel/voxel size of \spacingOneOne\ and \spacingOneOneOne. The results on the 3D dataset are stratified by organ categories: small, midsize, large and tubular (Fig.~\ref{fig:organs-groups} in Appendix).}
	\label{fig:compute-efficiency}
\end{figure*}

While accuracy is the primary focus of metric evaluation, \textit{computational efficiency} is also critical, as it enables rapid prototyping, real-time feedback, and reduced computational costs. For its assessment, we focused on \hd\ as it is the only metric consistently supported across all open-source tools, and measured execution times for \hd\ computation on the 2D and 3D datasets using unit pixel/voxel sizes. As shown in Fig.~\ref{fig:compute-efficiency}, the results revealed a substantial variability in median execution times, spanning two orders of magnitude. Overall, \googledeepmindtt\ followed by \monaitt\ emerged as the most, whereas \misevaltt\ as the least computationally efficient implementation.
\phantomsection
\section{Discussion}\label{sec:discussion}
\subsection{Boundary extraction}\label{sec:discussion-boudary-extraction}
Our conceptual analysis revealed two main paradigms in the boundary extraction step: one that considers all foreground elements as query points for distance computation, and another that uses only the boundary elements. A simple thought experiment including two segmentations with substantial overlap neatly illustrates a key difference: foreground-based calculations tend to yield lower \hdpercp, \masd\ and \assd\ values because overlapping regions produce zero distances, skewing the distribution toward smaller values and reducing sensitivity to boundary errors, whereas boundary-based calculations do not suffer from the same effect. It is therefore not surprising that distance-based metrics are often referred to as \textit{boundary-based}, as nine of the 11 open-source tools adopt boundary-based calculations.
\par
Nevertheless, among those nine tools, three different boundary extraction methods are used (Fig.~\ref{fig:boundary-extraction}). This variability most likely stems from the historical evolution of \hd, originally defined for point clouds. Unlike point clouds, segmentation masks are represented on fixed 2D/3D grids, raising the fundamental question: what precisely defines the boundary of a segmentation mask? The most intuitive answer -- the interface between foreground and background elements -- is challenging to be answered for the discretized grids. Most tools define the boundary as the outer ``peel'' of foreground elements (Fig.~\ref{fig:boundary-extraction-b} and \ref{fig:boundary-extraction-c}). In contrast, \googledeepmindtt\ and \pymiatt\ adopt an alternative approach: they shift the grid by half a pixel/voxel size to position the boundary directly at the interface between foreground and background elements (Fig.~\ref{fig:boundary-extraction-d}). This seems to be the closest approximation of the interface between the foreground and background that is achievable on the same-sized image grid.
\subsection{Mathematical definitions}\label{sec:discussion-mathematical-def}
Two types of differences in mathematical definitions were observed: \textit{inconsistencies} and \textit{adaptations to discretized calculation}. Inconsistencies were identified for \hdpercp\ and \masd. For \hdpercp, the ``Hausdorffian'' approach -- adopted by \metricsreloadedtt, \monaitt, \googledeepmindtt\ and \pymiatt\ -- computes the percentiles for each set of directed distances and then takes their maximum. In contrast, \segmetricstt, \medpy\ and \evaluatesegmentationtt\ aggregate over all distances directly, which can introduce bias when the cardinality of one distance set greatly exceeds the other. Furthermore, \plastimatchtt\ violates the fundamental property that \hd\ equals \hdpercp\ at the 100th percentile (\hd$\,{=}\,$\hdpercn{100}). For \masd, \animatt\ deviates from other implementations by taking the maximum of the two directed average distances instead of their mean. This divergence most likely stems from the absence of standardized definitions and naming conventions (cf.\ note under Table~\ref{tab:implementations-info}).
\par
The second type of differences arises from how analytical mathematical definitions are adapted to discretized calculations on image grids. This particularly impacts the aggregation of directed distances, such as percentile or average distance computations. Most implementations adopt a straightforward approach: they compute the $p$-th percentile directly from the sorted set of distances or average by dividing the sum of distances by the cardinality of the set, therefore implicitly assuming a uniform distribution of query points along the segmentation boundary. In contrast, the authors of \googledeepmindtt~\cite{Nikolov2021-DeepMind} (and \pymiatt, which inherits its code for \hdpercp\ and \nsd, but only for 3D) introduced a more sophisticated method. By recognizing that query points are in general unevenly distributed across boundaries (Fig.\ref{fig:boundary-extraction}), they argue that na\"ive aggregation can produce biased estimates. To address this, they assign each query point a corresponding boundary element size (i.e.\ a segment length in 2D or a surface area in 3D), and perform a weighted aggregation of percentiles and averages (Fig.~\ref{fig:equations}). This weighting is computed using an approximation of the marching cubes algorithm on discretized grids -- while not exact, it provides a substantially improved estimate of boundary contributions. Empirically, this effect is most pronounced on anisotropic grids (cf.\ \hdpercn{95} for \spacingHalfHalfTwo\ in Table\ref{tab:results-3D}).
\subsection{Edge case handling}\label{sec:discussion-edge-case-handling}
Scenarios where one or both segmentations are empty can, in principle, be handled outside the metric implementation; however, open-source tools should ensure consistent and predictable outputs. The most mathematically consistent convention is to return $\infty$\,mm for absolute metrics and $0\%$ for relative metrics when only one segmentation is empty, and respectively $0$\,mm and $100\%$ when both segmentations are empty~\cite{Maier-Hein2024-Metrics-Reloaded}. Providing clear warnings or informative messages can further aid in preventing misinterpretations in downstream analyses. Our analysis (Table~\ref{tab:edge-cases}) revealed widespread ambiguities in handling such edge cases, along with notable inconsistencies across the open-source tools. This is a critical issue that developers must address and end-users need to understand, as they ultimately bear responsibility for ensuring the validity of the reported results. For instance, when absolute metrics are undefined (i.e.\ one segmentation is empty), a common workaround involves replacing $\infty$ with either the maximum distance observed in the segmentation masks or a predefined penalizing factor to allow numerical analysis to proceed (Note~\ref{note:edge-case-handling} in Appendix). For an in-depth discussion of this issue, we refer readers to the \metricsreloadedit\ guidelines~\cite{Maier-Hein2024-Metrics-Reloaded}.
\subsection{Metric value variability}\label{sec:discussion-metric-value-variability}
The mean deviations for \hd\ are near zero across most tools\footnote{Except for \misevaltt, which does not account for pixel/voxel size (Fig.~\ref{fig:equations}).} (Tables~\ref{tab:results-2D} and \ref{tab:results-3D}), reflecting a broad agreement on its mathematical definition and its insensitivity to boundary extraction due to maximum-value aggregation. In contrast, other metrics exhibit considerable minimum and maximum deviations, and non-negligible mean differences (cf.\ the large spread of deviations in Figs.~\ref{fig:boxplots-2D} and \ref{fig:boxplots-3D} in Appendix). This is further substantiated by the statistical analysis (Figs.~\ref{fig:stat-testing-2D} and \ref{fig:stat-testing-3D} in Appendix), which reveals a concerning observation: for the exactly same set of segmentations, users may achieve statistically significant improvements in reported metric values simply by choosing a particular open-source tool. For large and tubular segmentations (Table~\ref{tab:results-per-organ-group} in Appendix), \plastimatchtt\ produced extreme outliers of up to $-115$\,mm due to averaging the two directed percentiles instead of taking their maximum, leading to overly optimistic results. Similarly, \medpytt\ and \segmetricstt\ showed over-optimistic performance by computing percentiles on the union of distance sets (Fig.~\ref{fig:equations}), which generally lowers metric values when combined with their boundary extraction methods. Conversely, \evaluatesegmentationtt, despite using the same aggregation approach, applies a boundary extraction method that biases the distribution toward larger distances, yielding overly pessimistic results. While \monaitt\ and \metricsreloadedtt\ appear highly accurate for isotropic grids, they exhibit outliers and reduced performance for anisotropic grids, attributable to the mathematical differences discussed earlier. This is perhaps best illustrated by the substantial deviations observed for \nsd, ranging from $-10.9$ to $51.6$\%pt and $-37.5$ to $7.0$\%pt for 2D and 3D datasets, respectively. Interestingly, \masd\ and \assd\ exhibit lower mean deviations across tools, as averaging over large distance sets dampens implementation variability. However, even modest average improvements of $0.3$\,mm in \masd\ or \assd\ are often regarded as meaningful in segmentation studies. Moreover, the observed deviation ranges (Min\,$|$\,Max) remain substantial, underscoring that virtually all distance-based metrics are vulnerable to implementation pitfalls. Notably, as \biou\ is a recently proposed metric, it is currently supported only by \metricsreloadedtt, which produces valid results solely for isotropic grids of unit size due to a flaw in its code (Note~\ref{note:evaluatesegmentation} in Appendix). This also underscores how critical it is for users to correctly specify the pixel/voxel size in metric computation pipelines, as neglecting this step can lead to errors of considerable magnitude.
\subsection{Computational efficiency}\label{sec:discussion-computational-efficiency}
Although accurate measurement should remain the top priority, computational efficiency is a practical consideration when selecting an open-source tool. We observed considerable differences in median execution times across all open-source tools (Fig.~\ref{fig:compute-efficiency}), with \googledeepmindtt\ and \monaitt\ being the most, and \misevaltt, \medpytt\, and \metricsreloadedtt\ the least computationally efficient tools. For the details on optimization strategies, please refer to Note~\ref{note:computational-efficiency} in Appendix.
\subsection{Limitations and future work}\label{sec:discussion-limitations-and-future-work}
This study focuses exclusively on distance-based metrics for evaluating image segmentation. While other metric categories exist, such as counting-based metrics (e.g.\ \dsc), their implementation on image grids is considerably less challenging. Indeed, although not reported, we also compared \dsc\ values across all 11 open-source tools and observed no discrepancies apart from negligible numerical differences at low decimal places. Several additional distance-based metrics exist, but we deliberately focused on the subset included in the \metricsreloadedit\ guidelines~\cite{Maier-Hein2024-Metrics-Reloaded}, and adopted their naming conventions to promote consistency and support unification efforts in the field of validation metrics. Although a systematic investigation of other metric groups -- including those beyond the segmentation domain -- would undoubtedly benefit the community, we leave this to future work, prioritizing here a concise yet representative analysis of distance-based metrics.
\par
The scope of this study is further limited to identifying and characterizing implementation pitfalls in existing open-source tools, which inherently confines our analysis to grid-based implementations. However, as suggested in our preliminary work~\cite{Podobnik2024HDilemma}, grid-based computations may not be ideal due to inherent discretization effects, such as the \textit{quantized distance effect}: a finite set of possible distance values exists and is determined by the pixel/voxel size and the arrangement of query points on the grid~\cite{Nikolov2021-DeepMind}. This effect is particularly pronounced when the pixel/voxel size and the \nsd/\biou\ margin parameter~$\tau$ are of similar magnitude. Computing metrics in the mesh domain may alleviate such limitations and improve accuracy, however, to maintain a focused scope, the development of such tools is left for future work.
\section{Conclusions}\label{sec:discussion-conclusions}
We would like to first acknowledge the authors of the 11 open-source tools~\cite{Commowick2018Anima, Taha2015-EvaluateSegmentation, Nikolov2021-DeepMind, Maier2019MedPy, Maier-Hein2024-Metrics-Reloaded, Muller2022-MISeval, Cardoso2022-MONAI, Zaffino2016-Plastimatch, Jungo2021pymia, Jia2024-SegMetrics, Lowekamp2013SimpleITK} for their valuable contributions. Despite the identified implementation pitfalls, open-source tools are the backbone of modern research and development, as they provide a more transparent and reproducible approach to method validation compared to closed-source, in-house solutions -- which may suffer from even more severe pitfalls, but remain out of reach of the community. Accordingly, the outcomes of this study should not be seen as a criticism but rather as a constructive step towards improving the implementation and use of validation metrics. Building on our conceptual and empirical analyses, we propose the following recommendations for distance-based metric computation:
\begin{itemize}
	\renewcommand\labelitemi{$\bullet$}
	\setlength\itemsep{0.25em}
	\item Consult metric selection guidelines, such as \metricsreloadedit~\cite{Maier-Hein2024-Metrics-Reloaded}, to identify representative metrics for the specific application. 
	\item Use an open-source tool for metric calculation, and explicitly report its name and version to ensure reproducibility. 
	\item Be mindful of edge cases -- do not rely solely on tool outputs. Use external checks to assign appropriate values when segmentations are empty. 
	\item Pay attention to the correct usage of units of measurement and pixel/voxel sizes. 
	\item For calculation on image grids, use \googledeepmindtt\ for \hd, \hdpercp, \masd\ and \assd\ due to its superior performance on both isotropic and anisotropic grids in comparison to other open-source tools, and exercise caution for \nsd\ and \biou\ due to implementation discrepancies. For applications requiring highly accurate distance-based metrics, particularly in the presence of discretization effects, consider mesh-based calculations.
\end{itemize}
\noindent
We hope that this breakdown of implementation pitfalls will raise community awareness and deepen the understanding of the computational principles underlying distance-based metrics. Our findings underscore the need for a more careful interpretation of segmentation results in both existing and future studies. Given the observed discrepancies, studies that relied on the 11 open-source tools, particularly the ones with major inconsistencies, may warrant re-evaluation for implementation errors. Looking ahead, we are optimistic that the identified pitfalls can be addressed in future updates and that novel tools for distance-based metric calculation can provide a robust solution to many of these challenges.
\section*{Research data}
The existing open-source tools for distance-based metric computation are publicly available through their corresponding repositories provided in Table~\ref{tab:implementations-info}. The datasets used in this study are publicly available under Creative Commons licenses as part of the \textit{INbreast dataset} (\url{https://www.kaggle.com/datasets/tommyngx/inbreast2012}; \texttt{CC~BY-NC~4.0})~\cite{Moreira2012INbreast} and \textit{HaN-Seg dataset} (\url{https://doi.org/10.5281/zenodo.7442914})~\cite{Podobnik2023-Han-Seg-Dataset}; \texttt{CC-NC-ND~4.0}).
\section*{Acknowledgments}
This study was supported by the Slovenian Research and Innovation Agency (ARIS) under projects No.\ J2-4453, J2-50067, J2-60042 and P2-0232, and by the European Union Horizon project ARTILLERY under grant agreement No.\ 101080983.
\bibliography{main}
\clearpage
\section*{Appendix}
\setcounter{secnumdepth}{2}
\renewcommand{\thesubsection}{A\arabic{subsection}}
\setcounter{subsection}{0}
\titleformat{\subsection}
	{\normalfont\bfseries}
	{Note~\thesubsection.}{0.5em}{}[]
\renewcommand{\thefigure}{A\arabic{figure}}
\setcounter{figure}{0}
\renewcommand{\thetable}{A\arabic{table}}
\setcounter{table}{0}
\def\Nan{\texttt{NaN}} 
\subsection{Metric nomenclature}\label{note:nomenclature}
As observed from the remarks in Table~\ref{tab:implementations-info}, there is a considerable ambiguity surrounding the \masd\ and \assd\ nomenclature. While we adopt the nomenclature from the \metricsreloadedit\ guidelines~\cite{Maier-Hein2024-Metrics-Reloaded}, it is important to acknowledge that various names are used for the same or similar metrics throughout the literature. To accommodate this, we took an inclusive approach, incorporating metrics from open-source tools with different naming conventions, as long as there was a strong resemblance in their mathematical definitions or a possibility that a less experienced user may confuse them with \masd\ or \assd. For example, \masd\ is often referred to simply as ASD by \evaluatesegmentationtt\ or average \hd\ by \plastimatchtt\ and \simpleitktt. It is worth noting that average \hd\ is inconsistently defined in the literature -- definitions using both the maximum~\cite{Taha2015-EvaluateSegmentation, Muller2022-MISeval} and mean~\cite{Aydin2021-AvgHD} of the directed average distances exist, sometimes even from the same authors. Some tools adopt a less common nomenclature, such as \texttt{ContourMeanDistance} and \texttt{SurfaceDistance} used by \animatt\ that we compare to \masd\ and \assd, respectively. Additionally, \monaitt\ provides a function to compute \assd\ with argument \texttt{symmetric}, which, when set to \texttt{False} by default, returns the average \textit{directed} distance that should not be confused with \masd. The remaining user options that some tools provide, such as the boundary extraction method in \monaitt, were left at their default settings, with the exception of the pixel/voxel size, which was explicitly specified.
\subsection{Similarities among open-source tools}
\label{note:similarities}
There are notable similarities between two pairs of open-source tools: \metricsreloadedtt\ and \monaitt, as well as \googledeepmindtt\ and \pymiatt. While \metricsreloadedtt\ and \monaitt\ share the underlying calculation principles, \metricsreloadedtt\ offers a more comprehensive set of metrics. Although \metricsreloadedtt\ is hosted within the \textit{Project MONAI} GitHub repository~\cite{Cardoso2022-MONAI}, both were developed independently without shared high-level code, and are therefore treated as distinct tools. 
On the other hand, \pymiatt\ directly adapts, with appropriate attribution, the implementations of \hd, \hdpercp\ and \nsd\ from \googledeepmindtt. For \masd, however, it uses the average \hd\ implementation from \simpleitktt. Notably, \pymiatt\ includes copied code snippets from \googledeepmindtt\ rather than using it as a dependency, meaning that updates to \googledeepmindtt\ are not reflected in \pymiatt. Additionally, their 2D implementations of \hd, \hdpercp\ and \nsd\ differ in boundary extraction methods, further justifying their treatment as distinct tools.
\subsection{Edge case handling}
\label{note:edge-case-handling}
Special attention must be given to aggregating results (i.e.\ in the form of tables or graphical presentations) in the presence of edge cases occupying \Nan\ and infinite values. Simply plotting these values using tools like \textit{Seaborn boxplot} (\textit{Seaborn} is a \texttt{Python} data visualization library) can lead to incorrect visualizations, as these values are ignored by default. Furthermore, if such edge cases exist, this suggests that the investigation may benefit from splitting the evaluation into the detection and segmentation part, ensuring objectivity and facilitating comparison to other studies. In some cases, the missing values still need to be replaced -- in line with the \textit{Metrics Reloaded} guidelines~\cite{Maier-Hein2024-Metrics-Reloaded}, we suggest replacing infinite values with the largest distance in the image (e.g.\ the diagonal) or the maximum distance computed across all evaluations, and documenting this decision.
\subsection{Implementation flaws}
\label{note:evaluatesegmentation}
Our analysis revealed several implementation flaws in the code of some open-source tools. \evaluatesegmentationtt\ employs optimization strategies~\cite{Taha2015-HD-Optimization} for \hd\ that are incorrectly applied to \hdpercp\ and \masd, leading to skewed distance distributions that produce non-deterministic results. \misevaltt\ ignores the pixel/voxel size when calculating distances between query points, therefore performing relatively accurately only for the isotropic grid of unit size, and \metricsreloadedtt\ exhibits a similar issue for \biou.
\subsection{Computational efficiency}
\label{note:computational-efficiency}
When observing the computational efficiency across different pixel/voxel sizes, the median execution time per case is the lowest for the small and midsize segmentations, which is expected because of the lower number of query points needed for distance calculation. Across different open-source tools, \googledeepmindtt\ was identified as the most computationally efficient, followed by \monaitt\ and \pymiatt. In comparison, solid median execution times can be observed for \plastimatchtt\ and \simpleitktt, while \animatt, \evaluatesegmentationtt, \medpytt, \metricsreloadedtt, \misevaltt\ and \segmetricstt\ are the least computationally efficient. The substantial accelerated computation times of \googledeepmindtt, \monaitt, \plastimatchtt\ and \pymiatt\ can be mostly attributed to the fact that they apply cropping to the bounding box before proceeding with boundary extraction and distance calculation. Given the fact that \hd, \hdpercp, \masd, \assd\ and \nsd\ all rely on the same two sets of directed distances, substantial optimization can be achieved by a simultaneous computation of multiple metrics. Indeed, \googledeepmindtt, \metricsreloadedtt, \plastimatchtt, \pymiatt, \segmetricstt\ and \simpleitktt\ calculate the required distances only once and reuse them for computation of multiple metrics, while \animatt, \evaluatesegmentationtt, \medpytt, \misevaltt\ and \monaitt\ recalculate the same distances for each individual metric.
\begin{figure*}[p]
	\centering
	\includegraphics[width=0.9\textwidth]{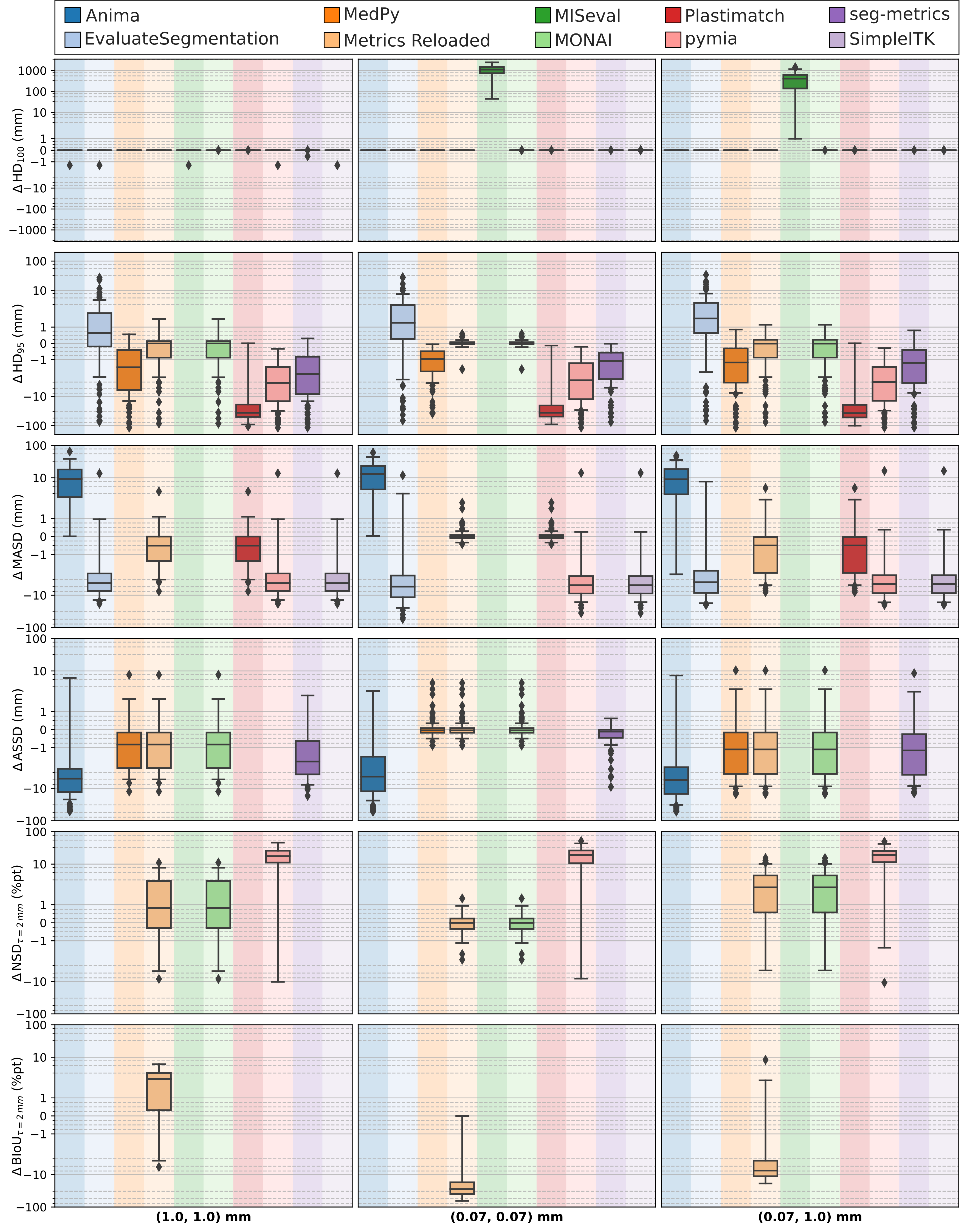}
	\vspace{0.25cm}
	\caption{The differences ($\Delta$) in the distance-based metrics on the 2D dataset, as obtained by each open-source tool against the reference \googledeepmindtt\ implementation on 80 pairs of 2D segmentation masks, reported for two isotropic and one anisotropic pixel size in the form of box-plots. Note that a linear scale is applied for values within the interval $[-1,1]$ and a symmetric logarithmic scale for values outside this interval to better visualize both smaller and larger differences. 
	}
	\label{fig:boxplots-2D}
\end{figure*}
\clearpage
\begin{figure*}[p]
	\centering
	\includegraphics[width=0.9\textwidth]{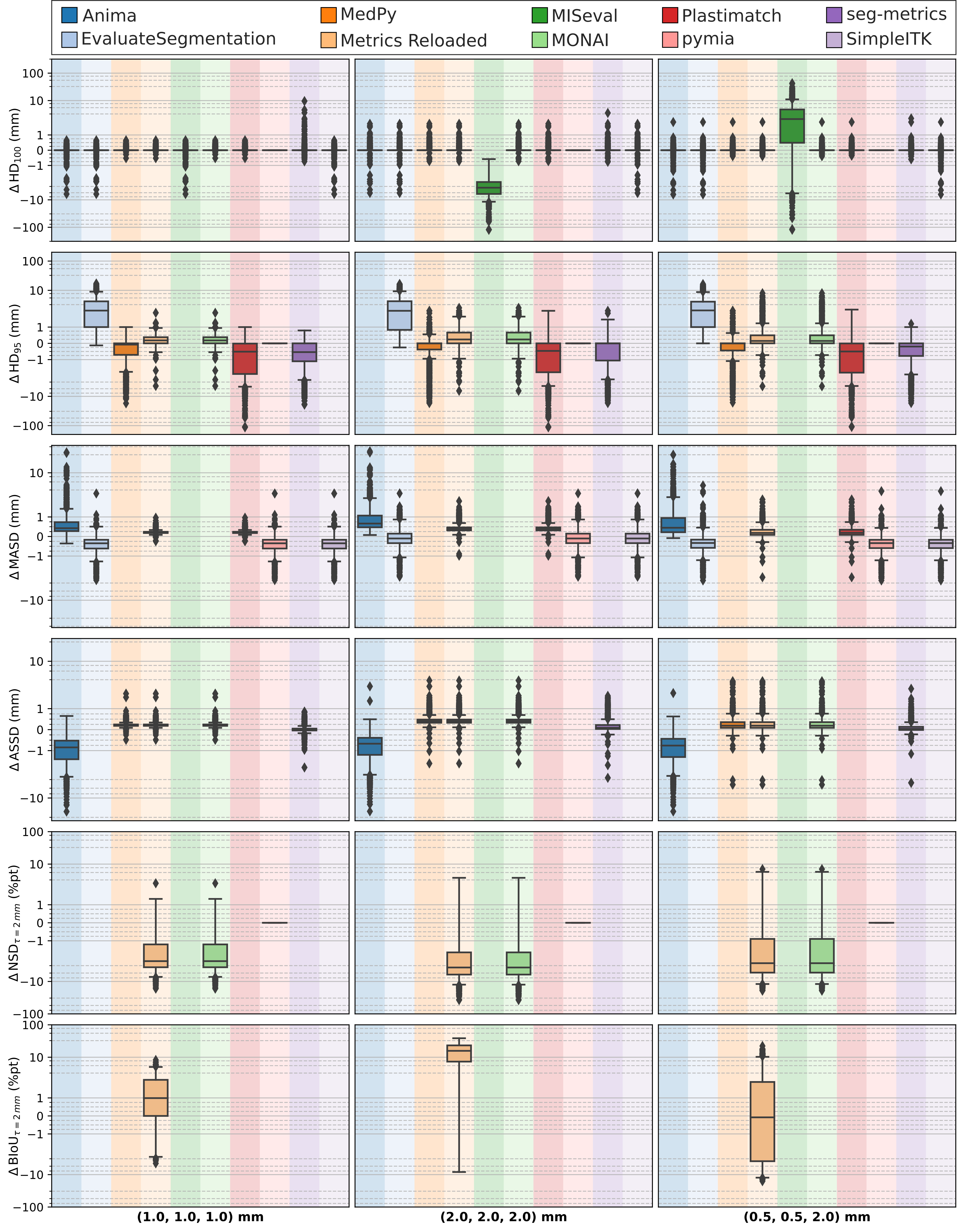}
	\vspace{0.25cm}
	\caption{The differences ($\Delta$) in the distance-based metrics on the 3D dataset, as obtained by each open-source tool against the reference \googledeepmindtt\ implementation on 1,559 pairs of 3D segmentation masks, reported for two isotropic and one anisotropic voxel size in the form of box-plots. Note that a linear scale is applied for values within the interval $[-1,1]$ and a symmetric logarithmic scale for values outside this interval to better visualize both smaller and larger differences. 
	}
	\label{fig:boxplots-3D}
\end{figure*}
\clearpage
\begin{figure*}[p]
	\centering
	\includegraphics[width=\textwidth]{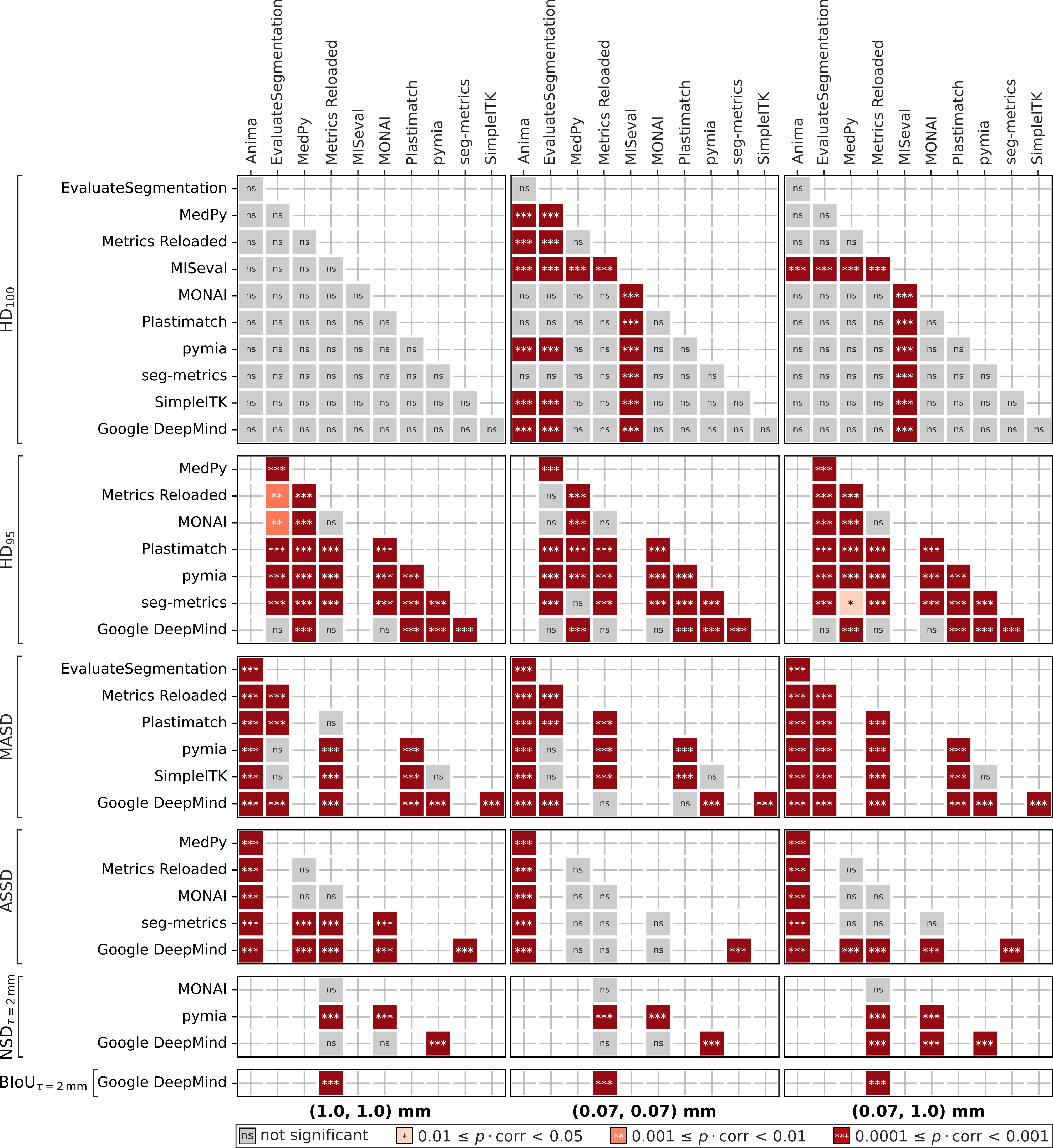}
	\vspace{0.25cm}
	\caption{Statistical testing of the open-source tools for distance-based metric computation on the 2D dataset including 80 pairs 2D segmentation masks, reported for two isotropic and one anisotropic voxel size. The value of \texttt{corr} is 990 and denotes the Bonferroni correction to adjust for the total number of comparisons and reduce the probability of type I error.}
	\label{fig:stat-testing-2D}
\end{figure*}
\clearpage
\begin{figure*}[p]
	\centering
	\includegraphics[width=\textwidth]{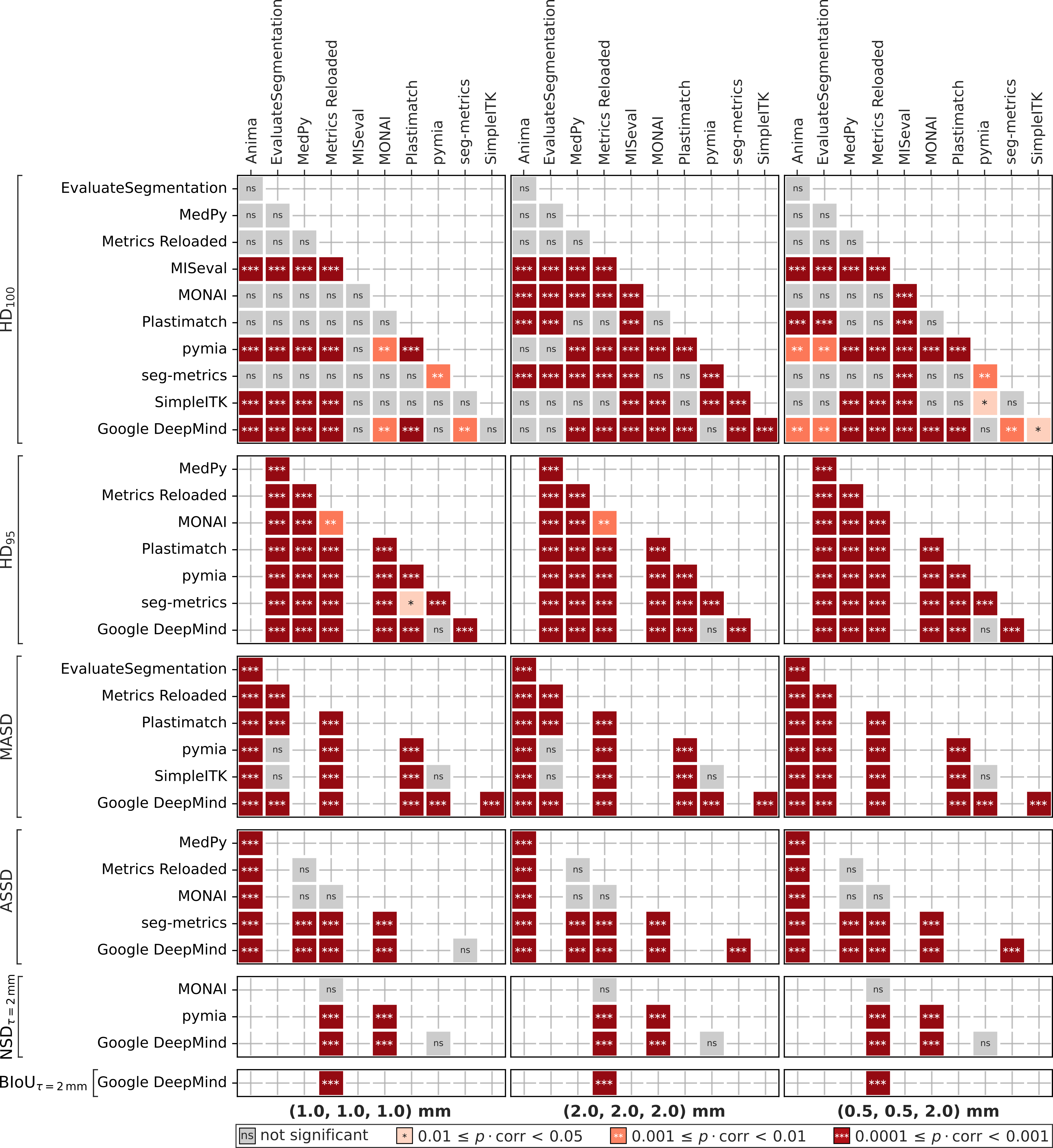}
	\vspace{0.25cm}
	\caption{Statistical testing of the open-source tools for distance-based metric computation on the 3D dataset including 1,559 pairs of 3D segmentation masks, reported for two isotropic and one anisotropic voxel size. The value of \texttt{corr} is 990 and denotes the Bonferroni correction to adjust for the total number of comparisons and reduce the probability of type I error.}
	\label{fig:stat-testing-3D}
\end{figure*}
\clearpage
\begin{figure*}[p]
	{\centering
	\includegraphics[width=\textwidth]{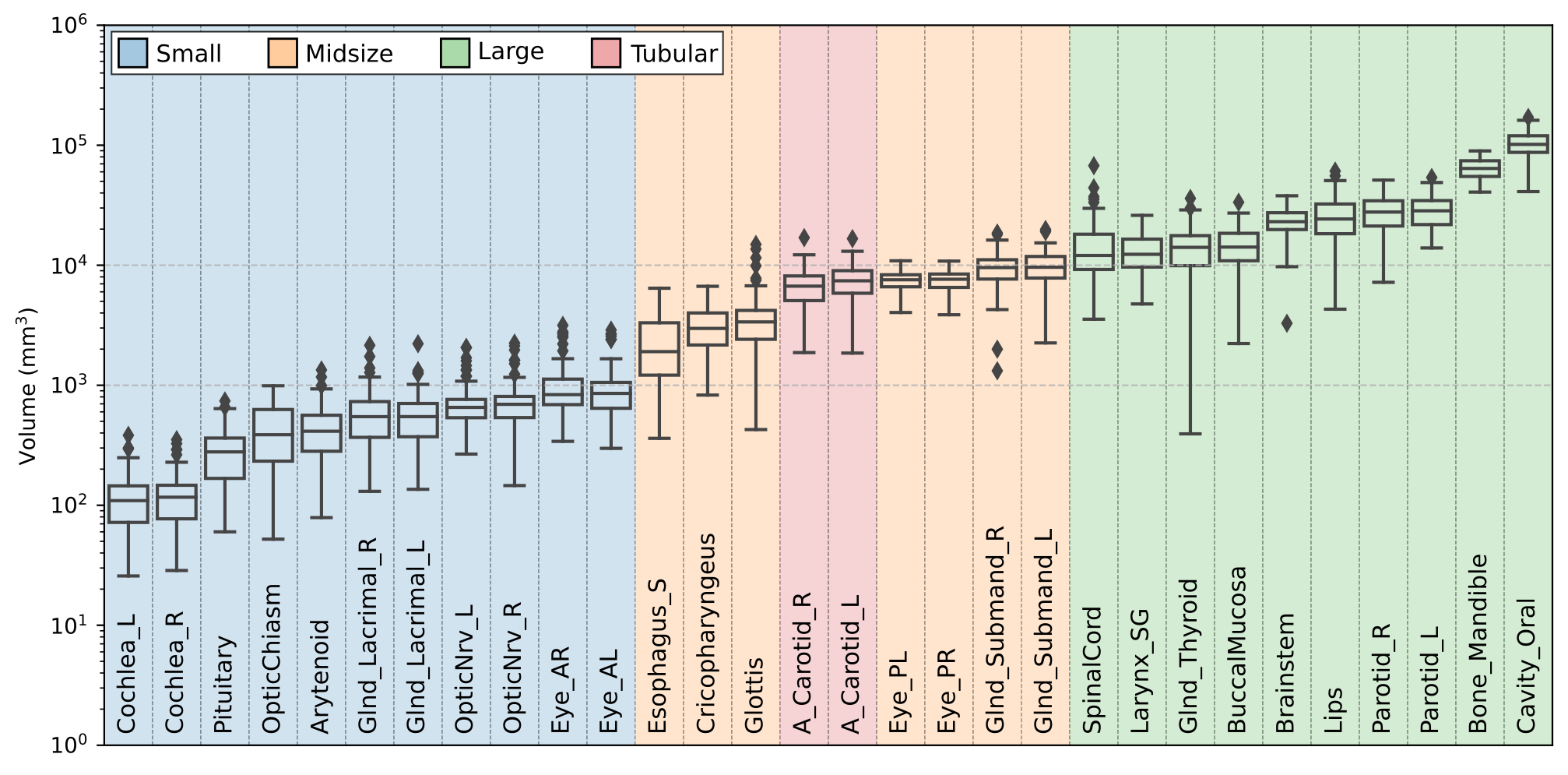}}
	\footnotesize{%
	\textbf{OAR labels} -- \texttt{A\_Carotid\_L}: left carotid artery, \texttt{A\_Carotid\_R}: right carotid artery, \texttt{Arytenoid}: arytenoids, \texttt{Bone\_Mandible}: mandible, \texttt{Brainstem}: brainstem, \texttt{BuccalMucosa}: buccal mucosa, \texttt{Cavity\_Oral}: oral cavity, \texttt{Cochlea\_L}: left cochlea, \texttt{Cochlea\_R}: right cochlea, \texttt{Cricopharyngeus}: cricopharyngeal inlet, \texttt{Esophagus\_S}: cervical esophagus, \texttt{Eye\_AL}: anterior segment of the left eyeball, \texttt{Eye\_AR}: anterior segment of the right eyeball, \texttt{Eye\_PL}: posterior segment of the left eyeball, \texttt{Eye\_PR}: posterior segment of the right eyeball, \texttt{Glnd\_Lacrimal\_L}: left lacrimal gland, \texttt{Glnd\_Lacrimal\_R}: right lacrimal gland, \texttt{Glnd\_Submand\_L}: left submandibular gland, \texttt{Glnd\_Submand\_R}: right submandibular gland, \texttt{Glnd\_Thyroid}: thyroid gland, \texttt{Glottis}: glottic larynx, \texttt{Larynx\_SG}: supraglottic larynx, \texttt{Lips}: lips, \texttt{Musc\_Constrict}: pharyngeal constrictor muscles, \texttt{OpticChiasm}: optic chiasm, \texttt{OpticNrv\_L}: left optic nerve, \texttt{OpticNrv\_R}: right optic nerve, \texttt{Parotid\_L}: left parotid gland, \texttt{Parotid\_R}: right parotid gland, \texttt{Pituitary}: pituitary gland, \texttt{SpinalCord}: spinal cord.}
	\vspace{0.5em}
	\vspace{0.25cm}
	\caption{The complete collection of $30$ organs-at risk (OARs) in the head and neck region of the \textit{HaN-Seg dataset}~\cite{Podobnik2023-Han-Seg-Dataset, Podobnik2024-vOARiability} that were used for distance-based metric computation within our experiment on real-world data, i.e.\ the 3D dataset, with their volumes $V$ shown in the form of box-plots. According to their median volume $\tilde{V}$, OARs with $\tilde{V}\,{\leq}\,10^3\,\textnormal{mm}^3$ were categorized as \textit{small}, OARs with $10^3\,{<}\,\tilde{V}\,{\leq}\,10^4\,\textnormal{mm}^3$ were categorized as \textit{midsize}, and OARs with $\tilde{V}\,{>}\,10^4\,\textnormal{mm}^3$ were categorized as \textit{large}. The left and right carotid arteries (\texttt{A\_Carotid\_L} and \texttt{A\_Carotid\_R}) were further categorized as \textit{tubular}, as such vessel-like structures often exhibit anomalies for specific metric values. The results of a detailed quantitative analysis of distance-based metrics computation are for different OAR categories presented in Table~\ref{tab:results-per-organ-group}.} 
	\label{fig:organs-groups}
\end{figure*}
\clearpage
\begin{table*}[p]
	\caption{The differences ($\Delta$) in the distance-based metrics on the 3D dataset, as computed by each open-source tool against the reference \googledeepmindtt\ implementation on 1,559 pairs of 3D segmentation masks, aggregated across all voxel sizes but stratified according to the organ-at-risk (OAR) category into small, midsize, large and tubular (Fig.~\ref{fig:organs-groups}), and reported as the range $\min{|}\max$ and mean $\pm$ standard deviation (SD).}
	\vspace{0.25cm}
	\label{tab:results-per-organ-group}
\vspace{-2pt}
\centering
\setlength{\tabcolsep}{-1.75pt}
\newcommand{\colS}{\hspace*{0.2cm}}
\newcolumntype{x}[1]{>{\centering\arraybackslash\hspace{0pt}}p{#1}}
\newcolumntype{C}[1]{>{\centering\let\newline\\\arraybackslash\hspace{0pt}}m{#1}}
\newcommand{\addFnote}[1]{\multicolumn{11}{x{0.985\textwidth}}{\footnotesize{#1}}}
\def\clen{1.8cm}
\newcommand{\sm}{\scalebox{0.5}[1.0]{\( - \)}}
\newcommand{\fm}{~\,\,}

\def\minmax{~~Min\,$|$\,Max}
\def\meanstd{Mean${\pm}$SD}
\begin{adjustbox}{width=\textwidth} 
\begin{footnotesize}
\begin{tabularx}{\linewidth}{l X C{\clen}C{\clen} X C{\clen}C{\clen} X C{\clen}C{\clen} X C{\clen}C{\clen}}
    \hline \noalign{\smallskip}
     & & \multicolumn{11}{c}{OAR category} \\
    \noalign{\smallskip} \cline{3-13} \noalign{\smallskip}
    Open-source tool & &
    \multicolumn{2}{c}{{Small}} & &
    \multicolumn{2}{c}{{Midsize}} & &
    \multicolumn{2}{c}{{Large}} & &
    \multicolumn{2}{c}{{Tubular}} \\
    %
\noalign{\smallskip} \midrule \noalign{\bigskip}
$\bf{\Delta\,HD}$ (mm) & \colS & \multicolumn{1}{c}{\minmax} & \multicolumn{1}{c}{\meanstd} & \colS & \minmax & \meanstd & \colS & \minmax & \meanstd & \colS & \minmax & \meanstd \\
\noalign{\smallskip} \hline \noalign{\smallskip}
\animatt                &  & ${\sm}\bf{0.00}\,{|}\,0.85$ & $0.01\,{\pm}\,0.05$ &  & ${\sm}2.25\,{|}\,1.53$ & $0.01\,{\pm}\,0.20$ &  & ${\sm}6.35\,{|}\,1.73$ & ${\sm}0.02\,{\pm}\,0.40$ &  & ${\sm}0.11\,{|}\,1.85$ & $0.02\,{\pm}\,0.14$ \\	
\evaluatesegmentationtt &  & ${\sm}\bf{0.00}\,{|}\,0.85$ & $0.01\,{\pm}\,0.05$ &  & ${\sm}2.25\,{|}\,1.53$ & $0.01\,{\pm}\,0.20$ &  & ${\sm}6.35\,{|}\,1.73$ & ${\sm}0.02\,{\pm}\,0.40$ &  & ${\sm}0.11\,{|}\,1.85$ & $0.02\,{\pm}\,0.14$ \\	
\medpytt                &  & $\bf{0.00}\,{|}\,0.85$ & $0.01\,{\pm}\,0.05$ &  & ${\sm}0.54\,{|}\,1.53$ & $0.03\,{\pm}\,0.13$ &  & ${\sm}0.72\,{|}\,1.73$ & $0.03\,{\pm}\,0.13$ &  & $\bf{0.00}\,{|}\,1.85$ & $0.02\,{\pm}\,0.14$ \\	
\metricsreloadedtt      &  & $\bf{0.00}\,{|}\,0.85$ & $0.01\,{\pm}\,0.05$ &  & ${\sm}0.54\,{|}\,1.53$ & $0.03\,{\pm}\,0.13$ &  & ${\sm}0.72\,{|}\,1.73$ & $0.03\,{\pm}\,0.13$ &  & $\bf{0.00}\,{|}\,1.85$ & $0.02\,{\pm}\,0.14$ \\	
\misevaltt              &  & ${\sm}15.8\,{|}\,21.6$ & $0.08\,{\pm}\,3.61$ &  & ${\sm}16.7\,{|}\,14.7$ & ${\sm}0.78\,{\pm}\,3.06$ &  & ${\sm}124\,{|}\,43.0$ & ${\sm}1.19\,{\pm}\,9.44$ &  & ${\sm}47.0\,{|}\,15.4$ & ${\sm}2.32\,{\pm}\,7.86$ \\	
\monaitt                &  & ${\sm}\bf{0.00}\,{|}\,0.85$ & $0.01\,{\pm}\,0.05$ &  & ${\sm}0.54\,{|}\,1.53$ & $0.03\,{\pm}\,0.13$ &  & ${\sm}0.72\,{|}\,1.73$ & $0.03\,{\pm}\,0.13$ &  & ${\sm}\bf{0.00}\,{|}\,1.85$ & $0.02\,{\pm}\,0.14$ \\	
\plastimatchtt          &  & ${\sm}\bf{0.00}\,{|}\,0.85$ & $0.01\,{\pm}\,0.05$ &  & ${\sm}0.54\,{|}\,1.53$ & $0.03\,{\pm}\,0.13$ &  & ${\sm}0.72\,{|}\,1.73$ & $0.03\,{\pm}\,0.13$ &  & ${\sm}\bf{0.00}\,{|}\,1.85$ & $0.02\,{\pm}\,0.14$ \\	
\pymiatt                &  & $\bf{0.00}\,{|}\,\bf{0.00}$ & $\bf{0.00}\,{\pm}\,\bf{0.00}$ &  & $\bf{0.00}\,{|}\,\bf{0.00}$ & $\bf{0.00}\,{\pm}\,\bf{0.00}$ &  & $\bf{0.00}\,{|}\,\bf{0.00}$ & $\bf{0.00}\,{\pm}\,\bf{0.00}$ &  & $\bf{0.00}\,{|}\,\bf{0.00}$ & $\bf{0.00}\,{\pm}\,\bf{0.00}$ \\	
\segmetricstt           &  & ${\sm}\bf{0.00}\,{|}\,0.85$ & $0.01\,{\pm}\,0.05$ &  & ${\sm}0.61\,{|}\,2.22$ & $0.03\,{\pm}\,0.18$ &  & ${\sm}0.76\,{|}\,9.57$ & $0.05\,{\pm}\,0.36$ &  & ${\sm}0.11\,{|}\,1.85$ & $0.02\,{\pm}\,0.14$ \\	
\simpleitktt            &  & $\bf{0.00}\,{|}\,0.85$ & $0.01\,{\pm}\,0.05$ &  & ${\sm}2.25\,{|}\,1.53$ & $0.01\,{\pm}\,0.20$ &  & ${\sm}6.35\,{|}\,1.73$ & ${\sm}0.02\,{\pm}\,0.40$ &  & ${\sm}0.11\,{|}\,1.85$ & $0.02\,{\pm}\,0.14$ \\	
 %
 %
 %
\noalign{\smallskip} \hline \noalign{\bigskip}
$\bf{\Delta\,HD_{95}}$ (mm) & \colS & \multicolumn{1}{c}{\minmax} & \multicolumn{1}{c}{\meanstd} & \colS & \minmax & \meanstd & \colS & \minmax & \meanstd & \colS & \minmax & \meanstd  \\
\noalign{\smallskip} \hline \noalign{\smallskip}
\evaluatesegmentationtt &  & ${\sm}\bf{0.00}\,{|}\,5.21$ & $1.13\,{\pm}\,0.87$ &  & ${\sm}0.25\,{|}\,6.44$ & $1.74\,{\pm}\,1.16$ &  & ${\sm}0.13\,{|}\,17.1$ & $4.81\,{\pm}\,2.80$ &  & $\bf{0.00}\,{|}\,15.2$ & $6.49\,{\pm}\,3.37$ \\	
\medpytt                &  & ${\sm}7.68\,{|}\,1.85$ & ${\sm}0.15\,{\pm}\,0.65$ &  & ${\sm}3.37\,{|}\,0.83$ & ${\sm}0.17\,{\pm}\,0.50$ &  & ${\sm}16.2\,{|}\,2.00$ & ${\sm}0.87\,{\pm}\,1.90$ &  & ${\sm}17.3\,{|}\,1.71$ & ${\sm}1.93\,{\pm}\,2.98$ \\	
\metricsreloadedtt      &  & ${\sm}2.00\,{|}\,1.94$ & $0.24\,{\pm}\,0.36$ &  & ${\sm}0.64\,{|}\,3.74$ & $0.26\,{\pm}\,0.39$ &  & ${\sm}4.52\,{|}\,8.00$ & $0.42\,{\pm}\,0.76$ &  & ${\sm}6.48\,{|}\,4.42$ & $0.50\,{\pm}\,0.95$ \\	
\monaitt                &  & ${\sm}2.00\,{|}\,1.94$ & $0.24\,{\pm}\,0.36$ &  & ${\sm}0.64\,{|}\,3.74$ & $0.26\,{\pm}\,0.39$ &  & ${\sm}4.52\,{|}\,8.00$ & $0.42\,{\pm}\,0.76$ &  & ${\sm}6.48\,{|}\,4.42$ & $0.50\,{\pm}\,0.95$ \\	
\plastimatchtt          &  & ${\sm}8.37\,{|}\,1.00$ & ${\sm}0.94\,{\pm}\,1.29$ &  & ${\sm}13.7\,{|}\,0.97$ & ${\sm}1.09\,{\pm}\,1.91$ &  & ${\sm}115\,{|}\,2.00$ & ${\sm}2.62\,{\pm}\,8.85$ &  & ${\sm}42.6\,{|}\,2.20$ & ${\sm}3.09\,{\pm}\,6.71$ \\	
\pymiatt                &  & $\bf{0.00}\,{|}\,\bf{0.00}$ & $\bf{0.00}\,{\pm}\,\bf{0.00}$ &  & $\bf{0.00}\,{|}\,\bf{0.00}$ & $\bf{0.00}\,{\pm}\,\bf{0.00}$ &  & $\bf{0.00}\,{|}\,\bf{0.00}$ & $\bf{0.00}\,{\pm}\,\bf{0.00}$ &  & $\bf{0.00}\,{|}\,\bf{0.00}$ & $\bf{0.00}\,{\pm}\,\bf{0.00}$ \\	
\segmetricstt           &  & ${\sm}8.66\,{|}\,1.85$ & ${\sm}0.28\,{\pm}\,0.75$ &  & ${\sm}4.00\,{|}\,0.83$ & ${\sm}0.50\,{\pm}\,0.75$ &  & ${\sm}14.5\,{|}\,2.00$ & ${\sm}1.33\,{\pm}\,2.08$ &  & ${\sm}19.2\,{|}\,1.16$ & ${\sm}2.47\,{\pm}\,3.37$ \\	
 %
\noalign{\smallskip} \hline \noalign{\bigskip}
$\bf{\Delta\,MASD}$ (mm) & \colS & \multicolumn{1}{c}{\minmax} & \multicolumn{1}{c}{\meanstd} & \colS & \minmax & \meanstd & \colS & \minmax & \meanstd & \colS & \minmax & \meanstd  \\
\noalign{\smallskip} \hline \noalign{\smallskip}
\animatt                &  & ${\sm}\bf{0.06}\,{|}\,5.28$ & $0.72\,{\pm}\,0.60$ &  & ${\sm}0.36\,{|}\,5.88$ & $0.70\,{\pm}\,0.78$ &  & ${\sm}\bf{0.05}\,{|}\,40.8$ & $1.10\,{\pm}\,2.67$ &  & $\bf{0.08}\,{|}\,17.2$ & $1.01\,{\pm}\,2.07$ \\	
\evaluatesegmentationtt &  & ${\sm}1.15\,{|}\,4.37$ & $0.03\,{\pm}\,0.40$ &  & ${\sm}1.61\,{|}\,\bf{1.22}$ & ${\sm}0.34\,{\pm}\,0.28$ &  & ${\sm}2.44\,{|}\,3.01$ & ${\sm}0.62\,{\pm}\,0.47$ &  & ${\sm}2.74\,{|}\,\bf{1.54}$ & ${\sm}\bf{0.16}\,{\pm}\,\bf{0.38}$ \\	
\metricsreloadedtt      &  & ${\sm}0.32\,{|}\,\bf{1.51}$ & $0.27\,{\pm}\,0.24$ &  & ${\sm}\bf{0.23}\,{|}\,1.29$ & $\bf{0.29}\,{\pm}\,\bf{0.18}$ &  & ${\sm}2.23\,{|}\,\bf{1.89}$ & $\bf{0.31}\,{\pm}\,\bf{0.22}$ &  & ${\sm}0.98\,{|}\,1.74$ & $0.28\,{\pm}\,0.20$ \\	
\plastimatchtt          &  & ${\sm}0.32\,{|}\,\bf{1.51}$ & $0.27\,{\pm}\,0.24$ &  & ${\sm}\bf{0.23}\,{|}\,1.30$ & $\bf{0.29}\,{\pm}\,\bf{0.18}$ &  & ${\sm}2.23\,{|}\,\bf{1.89}$ & $\bf{0.31}\,{\pm}\,\bf{0.22}$ &  & ${\sm}0.98\,{|}\,1.74$ & $0.28\,{\pm}\,0.20$ \\	
\pymiatt                &  & ${\sm}1.17\,{|}\,\bf{1.51}$ & $\bf{0.01}\,{\pm}\,\bf{0.37}$ &  & ${\sm}1.64\,{|}\,\bf{1.22}$ & ${\sm}0.34\,{\pm}\,0.28$ &  & ${\sm}2.44\,{|}\,3.01$ & ${\sm}0.63\,{\pm}\,0.48$ &  & ${\sm}2.74\,{|}\,\bf{1.54}$ & ${\sm}\bf{0.16}\,{\pm}\,\bf{0.38}$ \\	
\simpleitktt            &  & ${\sm}1.17\,{|}\,\bf{1.51}$ & $\bf{0.01}\,{\pm}\,\bf{0.37}$ &  & ${\sm}1.64\,{|}\,\bf{1.22}$ & ${\sm}0.34\,{\pm}\,0.28$ &  & ${\sm}2.44\,{|}\,3.01$ & ${\sm}0.63\,{\pm}\,0.48$ &  & ${\sm}2.74\,{|}\,\bf{1.54}$ & ${\sm}\bf{0.16}\,{\pm}\,\bf{0.38}$ \\	
 %
 %
\noalign{\smallskip} \hline \noalign{\bigskip}
$\bf{\Delta\,ASSD}$ (mm) & \colS & \multicolumn{1}{c}{\minmax} & \multicolumn{1}{c}{\meanstd} & \colS & \minmax & \meanstd & \colS & \minmax & \meanstd & \colS & \minmax & \meanstd  \\
\noalign{\smallskip} \hline \noalign{\smallskip}
\animatt                &  & ${\sm}7.07\,{|}\,\bf{0.49}$ & ${\sm}0.86\,{\pm}\,0.85$ &  & ${\sm}5.42\,{|}\,\bf{0.10}$ & ${\sm}0.89\,{\pm}\,0.75$ &  & ${\sm}13.3\,{|}\,2.16$ & ${\sm}1.37\,{\pm}\,1.21$ &  & ${\sm}22.8\,{|}\,{\sm}\bf{0.13}$ & ${\sm}1.31\,{\pm}\,2.92$ \\	
\medpytt                &  & ${\sm}\bf{0.24}\,{|}\,1.69$ & $0.30\,{\pm}\,0.25$ &  & ${\sm}\bf{0.22}\,{|}\,1.86$ & $0.30\,{\pm}\,0.19$ &  & ${\sm}4.43\,{|}\,3.09$ & $0.32\,{\pm}\,0.29$ &  & ${\sm}\bf{1.61}\,{|}\,2.97$ & $0.29\,{\pm}\,0.30$ \\	
\metricsreloadedtt      &  & ${\sm}\bf{0.24}\,{|}\,1.69$ & $0.30\,{\pm}\,0.25$ &  & ${\sm}\bf{0.22}\,{|}\,1.86$ & $0.30\,{\pm}\,0.19$ &  & ${\sm}4.43\,{|}\,3.09$ & $0.32\,{\pm}\,0.29$ &  & ${\sm}\bf{1.61}\,{|}\,2.97$ & $0.29\,{\pm}\,0.30$ \\	
\monaitt                &  & ${\sm}\bf{0.24}\,{|}\,1.69$ & $0.30\,{\pm}\,0.25$ &  & ${\sm}\bf{0.22}\,{|}\,1.86$ & $0.30\,{\pm}\,0.19$ &  & ${\sm}4.43\,{|}\,3.09$ & $0.32\,{\pm}\,0.29$ &  & ${\sm}\bf{1.61}\,{|}\,2.97$ & $0.29\,{\pm}\,0.30$ \\	
\segmetricstt           &  & ${\sm}0.35\,{|}\,1.61$ & $\bf{0.18}\,{\pm}\,\bf{0.28}$ &  & ${\sm}0.54\,{|}\,1.51$ & $\bf{0.06}\,{\pm}\,\bf{0.17}$ &  & ${\sm}\bf{3.97}\,{|}\,\bf{1.95}$ & $\bf{0.02}\,{\pm}\,\bf{0.23}$ &  & ${\sm}2.93\,{|}\,1.48$ & $\bf{0.05}\,{\pm}\,\bf{0.29}$ \\
 %
 %
\noalign{\smallskip} \hline \noalign{\bigskip}
$\bf{\Delta\,NSD_{\,\tau=2\,mm}}$ (\%pt) & \colS & \multicolumn{1}{c}{\minmax} & \multicolumn{1}{c}{\meanstd} & \colS & \minmax & \meanstd & \colS & \minmax & \meanstd & \colS & \minmax & \meanstd  \\
\noalign{\smallskip} \hline \noalign{\smallskip}
\metricsreloadedtt      &  & ${\sm}37.5\,{|}\,5.41$ & ${\sm}4.34\,{\pm}\,4.67$ &  & ${\sm}20.6\,{|}\,5.50$ & ${\sm}3.18\,{\pm}\,2.84$ &  & ${\sm}16.5\,{|}\,7.00$ & ${\sm}3.60\,{\pm}\,2.81$ &  & ${\sm}13.7\,{|}\,3.77$ & ${\sm}1.82\,{\pm}\,1.76$ \\	
\monaitt                &  & ${\sm}37.5\,{|}\,5.41$ & ${\sm}4.34\,{\pm}\,4.67$ &  & ${\sm}20.6\,{|}\,5.50$ & ${\sm}3.18\,{\pm}\,2.84$ &  & ${\sm}16.5\,{|}\,7.00$ & ${\sm}3.60\,{\pm}\,2.81$ &  & ${\sm}13.7\,{|}\,3.77$ & ${\sm}1.82\,{\pm}\,1.76$ \\	
\pymiatt                &  & $\bf{0.00}\,{|}\,\bf{0.00}$ & $\bf{0.00}\,{\pm}\,\bf{0.00}$ &  & $\bf{0.00}\,{|}\,\bf{0.00}$ & $\bf{0.00}\,{\pm}\,\bf{0.00}$ &  & $\bf{0.00}\,{|}\,\bf{0.00}$ & $\bf{0.00}\,{\pm}\,\bf{0.00}$ &  & $\bf{0.00}\,{|}\,\bf{0.00}$ & $\bf{0.00}\,{\pm}\,\bf{0.00}$ \\	
 %
 %
\noalign{\smallskip} \hline \noalign{\bigskip}
$\bf{\Delta\,BIoU_{\,\tau=2\,mm}}$ (\%pt) & \colS & \multicolumn{1}{c}{\minmax} & \multicolumn{1}{c}{\meanstd} & \colS & \minmax & \meanstd & \colS & \minmax & \meanstd  & \colS & \minmax & \meanstd \\
\noalign{\smallskip} \hline \noalign{\smallskip}
\metricsreloadedtt      &  & ${\sm}\bf{11.2}\,{|}\,\bf{31.5}$ & $\bf{2.96}\,{\pm}\,\bf{5.66}$ &  & ${\sm}\bf{12.1}\,{|}\,\bf{38.5}$ & $\bf{7.25}\,{\pm}\,\bf{10.7}$ &  & ${\sm}\bf{15.9}\,{|}\,\bf{35.1}$ & $\bf{6.81}\,{\pm}\,\bf{11.2}$ &  & ${\sm}\bf{12.2}\,{|}\,\bf{20.8}$ & $\bf{2.47}\,{\pm}\,\bf{8.40}$ \\
 %
 %
\noalign{\smallskip} \hline
\end{tabularx}
\end{footnotesize}
\end{adjustbox}  
\end{table*}
\clearpage
\end{document}